\documentclass[11pt, letterpaper]{template}
\usepackage[authoryear, round]{natbib}

\usepackage{subfigure}
\usepackage{subcaption}
\usepackage{calc}
\usepackage[export]{adjustbox}
\usepackage[version=4]{mhchem} 
\usepackage{multirow}
\usepackage[T1]{fontenc}
\usepackage{lmodern}

\usepackage{longtable}
\usepackage{framed}
\definecolor{shadecolor}{RGB}{248,248,248}

\usepackage[capitalize,noabbrev]{cleveref}

\usepackage[dvipsnames]{xcolor}         
\hypersetup{
    colorlinks=true,
    linkcolor=Cyan,
    citecolor=Green,
    filecolor=magenta,
    urlcolor=magenta,
}

\theoremstyle{plain}

\theoremstyle{definition}

\theoremstyle{remark}

\usepackage[textsize=tiny]{todonotes}

\usepackage{listings}
\usepackage[most]{tcolorbox}
\definecolor{NavyBlue}{rgb}{0.0, 0.0, 0.5}
\definecolor{DarkerNavy}{rgb}{0.0, 0.05, 0.3}
\definecolor{DeepNavy}{rgb}{0.0, 0.0, 0.4}
\definecolor{DustyBlue}{rgb}{0.2, 0.3, 0.4}
\tcbset{
  promptbox/.style={
    top=5pt,
    bottom=5pt,
    colback=lightgray!20,
    colframe=black,
    colbacktitle=DustyBlue,
    enhanced,
    center,
    attach boxed title to top center={yshift=-0.1in,xshift=0.0in},
    boxed title style={boxrule=0pt,colframe=white,},
    before upper={\parindent0pt\setlength{\parskip}{2pt}},
  }
}
\newtcolorbox{promptbox}[2][]{promptbox, title=#2,#1}

\newtcbtheorem[number within=section]{question}{Question}{
    top=10pt,
    colback=lightgray!20,
    colframe=Black,
    colbacktitle=DustyBlue,
    enhanced,
    center,
    attach boxed title to top center={yshift=-0.1in,xshift=0.0in},
    boxed title style={boxrule=0pt,colframe=white,},
    breakable
}{qn}

\newtcbtheorem[number within=section]{prompt}{Prompt}{
    top=10pt,
    colback=lightgray!20,
    colframe=Black,
    colbacktitle=Cerulean,
    enhanced,
    center,
    attach boxed title to top center={yshift=-0.1in,xshift=0.0in},
    boxed title style={boxrule=0pt,colframe=white,},
    breakable
}{pp}

\newtcbtheorem[number within=section]{case}{Case}{
    top=10pt,
    colback=lightgray!20,
    colframe=Black,
    colbacktitle=Tan,
    enhanced,
    center,
    attach boxed title to top center={yshift=-0.1in,xshift=0.0in},
    boxed title style={boxrule=0pt,colframe=white,},
    breakable
}{cs}

\usepackage{xspace}
\newcommand{\name}{\texttt{ATLAS}\xspace}

\newcommand{\tightlist}{%
\setlength{\itemsep}{0pt}\setlength{\parskip}{0pt}}

\usepackage{pifont}

\title{\name: A High-Difficulty, Multidisciplinary Benchmark for Frontier Scientific Reasoning}

\author[*]{ATLAS Teams\footnote{Detailed contributor list in \Cref{app:contributors}.}}
\affil[*]{Shanghai AI Laboratory}

\begin{document}

\begin{abstract}

The rapid advancement of Large Language Models (LLMs) has led to performance saturation on many established benchmarks, questioning their ability to distinguish frontier models. Concurrently, existing high-difficulty benchmarks often suffer from narrow disciplinary focus, oversimplified answer formats, and vulnerability to data contamination, creating a fidelity gap with real-world scientific inquiry. To address these challenges, we introduce
\textbf{\name} (\textbf{A}GI-Oriented  \textbf{T}estbed for \textbf{L}ogical \textbf{A}pplication in \textbf{S}cience), a large-scale, high-difficulty, and cross-disciplinary evaluation suite composed of approximately 800 original problems. Developed by domain experts (PhD-level and above), \name spans seven core scientific fields: mathematics, physics, chemistry, biology, computer science, earth science, and materials science. Its key features include: (1) \textbf{High Originality and Contamination Resistance}, with all questions newly created or substantially adapted to prevent test data leakage; (2) \textbf{Cross-Disciplinary Focus}, designed to assess models' ability to integrate knowledge and reason across scientific domains; (3) \textbf{High-Fidelity Answers}, prioritizing complex, open-ended answers involving multi-step reasoning and LaTeX-formatted expressions over simple multiple-choice questions; and (4) \textbf{Rigorous Quality Control}, employing a multi-stage process of expert peer review and adversarial testing to ensure question difficulty, scientific value, and correctness. We also propose a robust evaluation paradigm using a panel of LLM judges for automated, nuanced assessment of complex answers. Preliminary results on leading models demonstrate \name's effectiveness in differentiating their advanced scientific reasoning capabilities. We plan to develop \name into a long-term, open, community-driven platform to provide a reliable ``ruler'' for progress toward Artificial General Intelligence. The project is released at: \url{https://github.com/open-compass/ATLAS}
\end{abstract}

\maketitle

\begin{figure*}[hbp!]

    \centering
    \begin{minipage}[c]{0.5\textwidth}
        \centering
        \includegraphics[width=\linewidth]{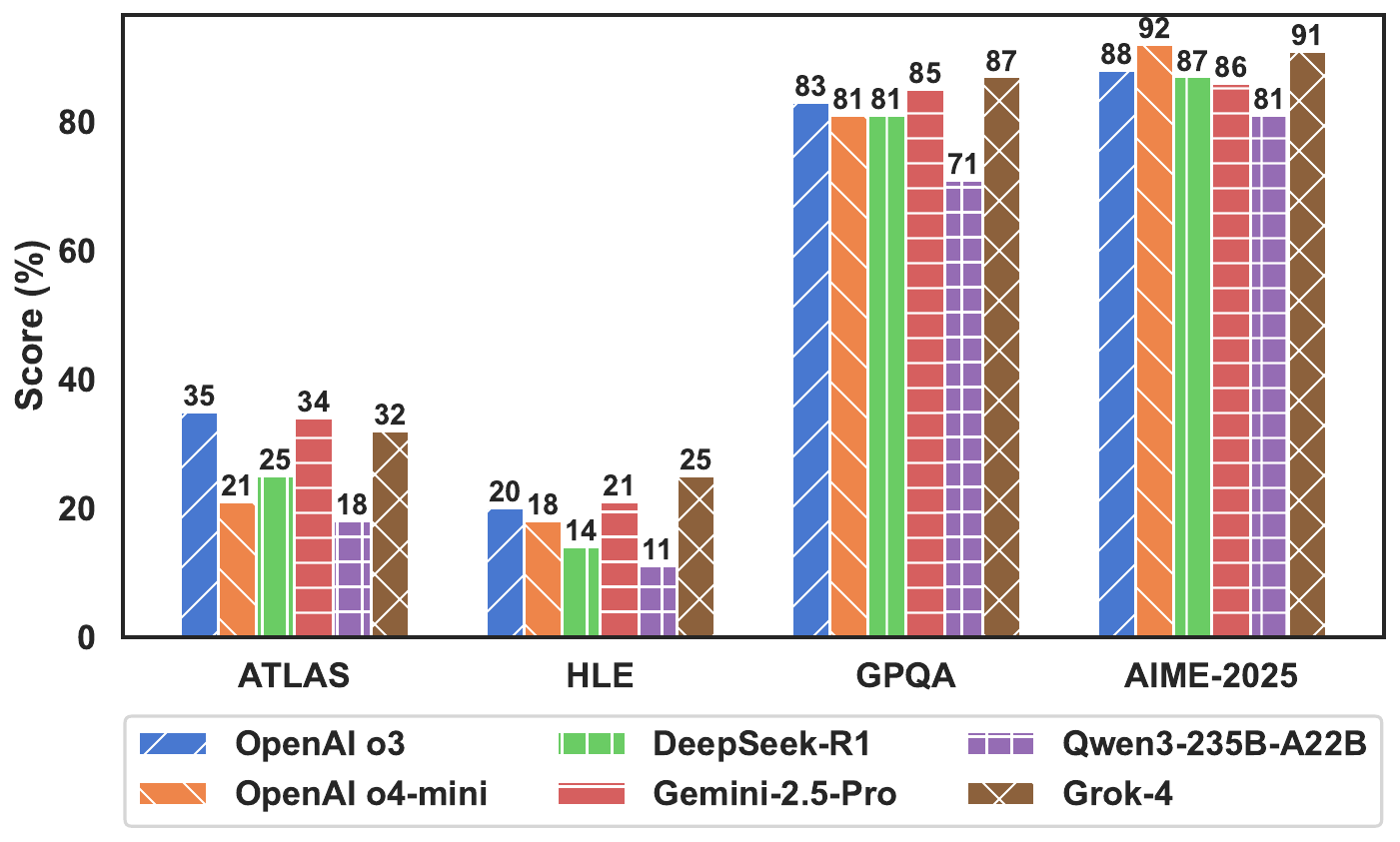}
        \caption{Reasoning LLMs performance comparison between \name and other commonly used reasoning benchmarks.}
        \label{fig:score_compare}
    \end{minipage}
    \hspace{5pt}
    \begin{minipage}[c]{0.4\textwidth}
        \centering
        \includegraphics[width=.9\linewidth]{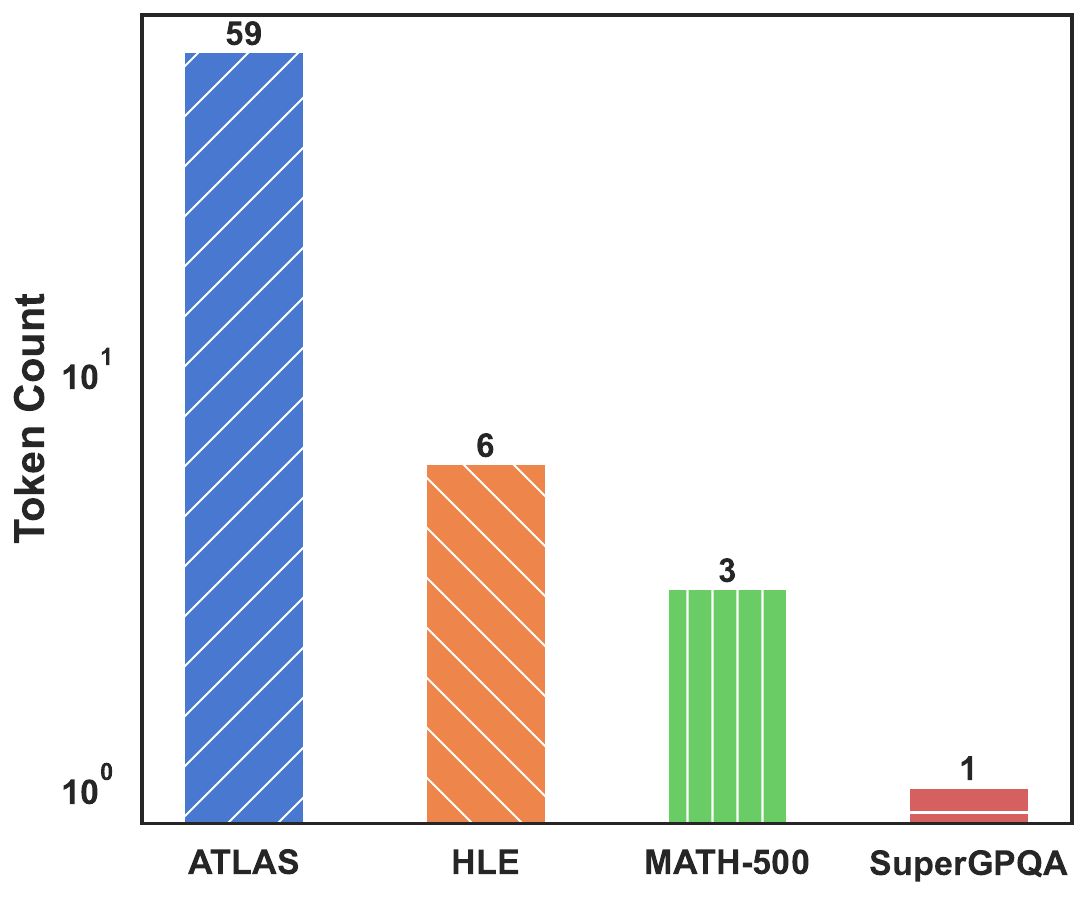}
        \caption{Average final answer token length for mainstream reasoning datasets.}
        \label{fig:answer_len}
    \end{minipage}
\end{figure*}
\newpage

\begin{figure*}[ht!]
    \centering
    \includegraphics[scale=0.3]{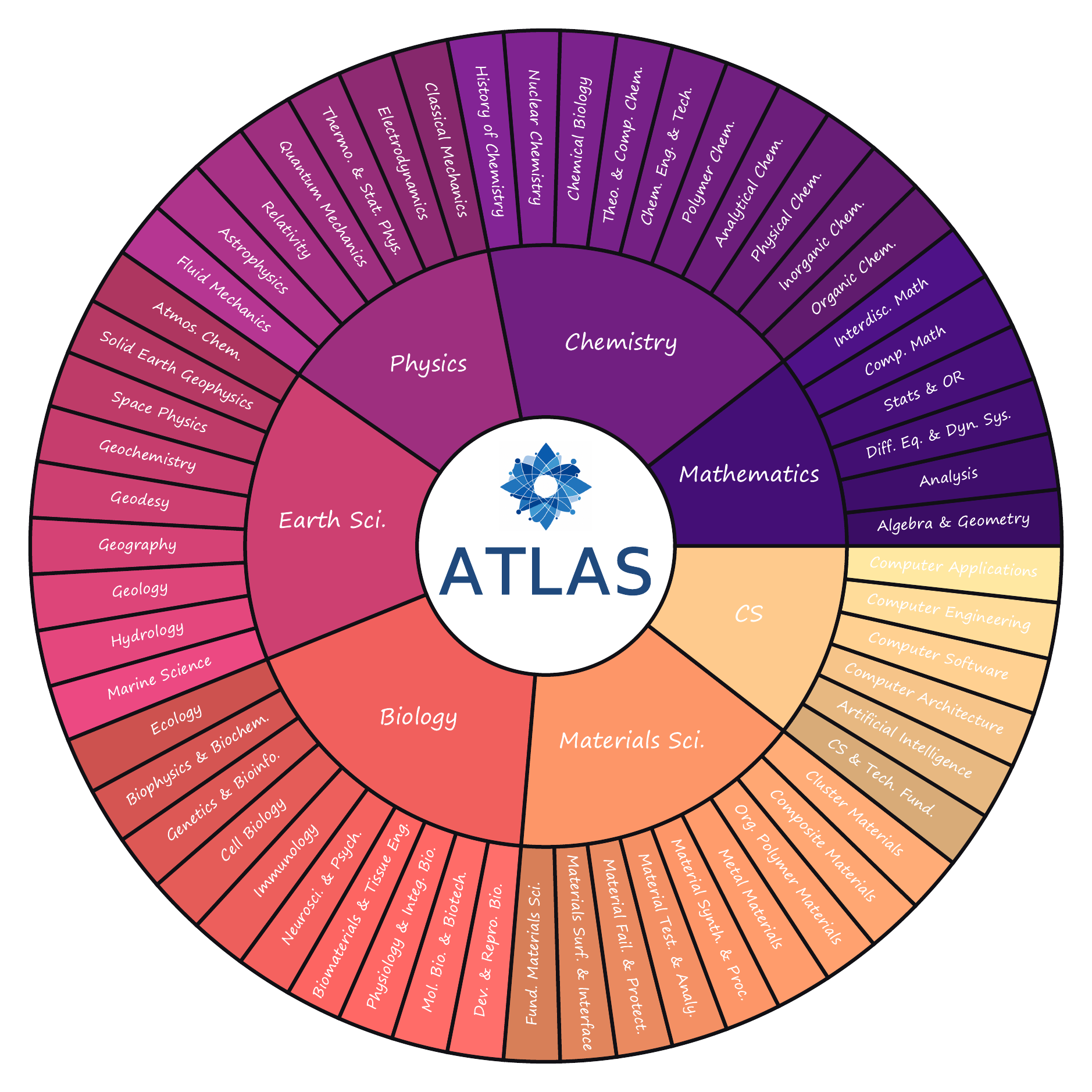}
    \caption{Overview of \name, which contains 7 stem subjects and 57 corresponding sub-fields.}
    \label{fig:main_figure}
\end{figure*}

\section{Introduction}

\subsection{Benchmark Saturation Phenomenon}
In recent years, the advancement of Large Language Models (LLMs) has been remarkable, with their performance on various natural language processing tasks approaching or even surpassing human levels. 
However, this rapid progress has resulted in a significant issue: the ``benchmark saturation'' of standardized evaluation sets. Many benchmarks previously regarded as ``gold standards'', such as MMLU, are now easily surpassed by state-of-the-art models with accuracies exceeding 90\%~\citep{yue2025hle, hendrycks2020mmlu}. 
This phenomenon reduces the effectiveness of these benchmarks in distinguishing the true capabilities of different models, particularly the subtle differences among cutting-edge models. 
A prominent example is the MATH dataset; upon its release in 2021, the leading model achieved a score of less than 10\%. Within just three years, top models have achieved scores over 90\%~\citep{hendrycksmath2021}. 
This situation underscores the urgent need for a new generation of more challenging evaluation tools to accurately assess and propel the continuous development of AI capabilities.

\subsection{Evaluation Needs for Frontier Scientific Reasoning}
The next substantial breakthrough in artificial intelligence is anticipated to involve solving complex, high-value real-world problems, with scientific discovery as a central focus~\citep{WangFD0HLCLKDAB23}. 
AI for Science (AI4S) seeks to expedite the scientific research process through AI, necessitating models to have not only a robust knowledge base but also advanced, multi-step, and interdisciplinary reasoning skills~\citep{ReddyS25,abs-2503-05822,abs-2502-18864}. 
To guide and assess the development of models in this strategic direction, it is essential to construct evaluation benchmarks that specifically test these capabilities. 
\name is developed for this purpose, aiming to serve as a ``touchstone'' for the AI4S domain, accurately reflecting the scientific reasoning abilities of models.

\subsection{Limitations of Existing High-Difficulty Benchmarks}
To tackle the challenge of benchmark saturation, the research community has developed several high-difficulty evaluation sets. 
While these initiatives have made significant contributions, they also exhibit limitations. 
Some benchmarks, despite their difficulty, are overly narrow in scope. For instance, MATH~\citep{hendrycksmath2021}, MathBench~\citep{liu2024mathbench} and OlympiadBench~\citep{he-etal-2024-olympiadbench} predominantly focus on mathematics or physics competition problems, hindering comprehensive evaluation of a model's integrated reasoning capabilities across diverse scientific domains. 
Conversely, benchmarks with broader coverage, such as Humanity's Last Exam (HLE)~\citep{yue2025hle} and SuperGPQA~\citep{du2025supergpqa}, while extremely challenging, are designed to assess general academic knowledge and are not specifically tailored for the deep, integrated scientific reasoning essential in the AI4S domain. Moreover, many existing benchmarks (e.g., AGIEval~\citep{zhong-etal-2024-agieval}, OlympiadBench~\citep{he-etal-2024-olympiadbench})  derive problems from public exam or competition question banks, posing a persistent risk of data contamination. 
Models might score highly by having encountered similar or identical problems during training, reflecting memorization rather than authentic reasoning abilities. 
One of the primary objectives of \name is to fundamentally resolve this issue through a rigorous original problem-setting approach.

Another limitation is that, for ease of verification, much of the existing work converts problems into multiple-choice questions and simple symbolic expressions~\citep{hendrycksmath2021,rein2023gpqa,du2025supergpqa,yue2025hle}. 
This method has resulted in a disconnect between benchmarks and real-world questions, particularly in scientific domains~\citep{abs-2505-08253}. 
In an era of rapid LLM capability expansion, benchmarks should not be confined to easily verifiable problems. 
\name is designed to preserve real-world problems and solutions, encompassing multiple sub-questions and complex natural and symbolic language expressions, to provide a more realistic and effective evaluation of a model's scientific capabilities.
To address the evaluation bottleneck, we propose an effective,  transferable and scalable LRM-as-Judge-based~\citep{vicuna2023,ZhengC00WZL0LXZ23} evaluation workflow, wherein Large Reasoning Models serve as judge models.
We anticipate that as model capabilities advance, the effectiveness of our workflow will be further improved. Concurrently, \name is poised to significantly contribute to the development of LRM-as-Judge.

\subsection{Our Contributions}
This study aims to overcome the aforementioned challenges by constructing \name. 
Its core contributions can be summarized in the following four points:
\begin{enumerate}
        \item \textbf{\name:} We release a new, highly challenging evaluation benchmark containing approximately 800 expert-created original problems. The benchmark focuses on multidisciplinary scientific reasoning, with a target difficulty set to a pass rate of less than 20\% for current state-of-the-art models, to effectively measure the true capabilities of frontier models. \name preserves real-world problems and solutions for realistic and effective evaluation of scientific capabilities.
    \item \textbf{A Rigorous, Contamination-Resistant Construction Pipeline:} We detail an innovative, multi-stage data generation and validation process. This process (as shown in \Cref{fig:workflow}) deeply integrates the wisdom of human experts with the adversarial testing of large models, ensuring the originality, high quality, and high difficulty of the problems from the source.

    \item \textbf{A Transferable and Scalable Evaluation Workflow: } We present a streamlined and scalable evaluation workflow that utilizes LRM-as-Judge paradigm. This approach facilitates efficient and automated evaluations, allowing researchers and practitioners to assess the reasoning capabilities of their models and conduct reinforcement learning on real-world benchmarks.
    \item \textbf{A Sustainable Evaluation Platform:} The release of \name is the first step in our long-term plan. Our ultimate goal is to build a community-driven collaborative platform that continuously generates and releases high-quality evaluation sets, thereby enabling the long-term, dynamic tracking of progress toward Artificial General Intelligence (AGI).
\end{enumerate}


\section{Related Work}
The swift advancement of LLMs necessitates a concurrent evolution in evaluation benchmarks, driving them towards increased difficulty, breadth, and methodological rigor. 
We contextualize our research by examining three significant trends: 
the transition from broad-coverage to human-centric assessments, 
the emergence of frontier-difficulty reasoning benchmarks, and 
the creation of specialized STEM evaluations.
Finally, we also discuss the related work of LLM-as-Judge, which is highly related to the evaluation work of \name.

\subsection{From Broad Coverage to Human-Centric Benchmarks}

 Initial comprehensive benchmarks like \textbf{MMLU} \citep{hendrycks2020mmlu}, with its multiple-choice questions across 47 subjects, have become "saturated" by state-of-the-art models, reducing their ability to differentiate frontier capabilities \citep{yue2025hle}. In response, human-centric benchmarks emerged, drawing questions from high-stakes standardized tests to ensure quality and relevance to human cognition. For instance, \textbf{AGIEval} \citep{zhong-etal-2024-agieval} uses questions from exams like the SAT and Gaokao, and \textbf{C-Eval} \citep{huang2023ceval} focuses on Chinese academic disciplines. While valuable, these benchmarks are constrained by the difficulty of their source material and face a significant, unavoidable risk of data contamination, as test questions are often public \citep{brown2020language, li2024opensource}. Our work directly mitigates these issues through expert-authored, original problems.

\subsection{The Rise of Frontier-Difficulty Reasoning Benchmarks}

 To address the limitations of existing tests, a new generation of benchmarks aims to create problems at the frontier of machine capabilities, emphasizing originality and resistance to search engine-based solutions. \textbf{GPQA} \citep{rein2023gpqa} exemplifies this with graduate-level questions whose creation process by multiple domain experts makes them demonstrably ``Google-proof”: experts achieved 65\% accuracy while skilled non-experts with web access only reached 34\% \citep{bowman2021benchmarking}. Similarly, \textbf{Humanity's Last Exam (HLE)} \citep{yue2025hle} employs nearly 1,000 experts and uses state-of-the-art models as an adversarial filter to ensure its 2,500 questions are challenging even for top models like Gemini 2.5 Pro (21.64\% accuracy). \name adopts the rigorous methodological principles of GPQA and HLE—such as expert-driven creation and adversarial filtering \citep{le2020adversarial, kiela2021dynabench}—but narrows the focus from general knowledge to the specific, high-value domain of AI for Science.


 A parallel research thrust has created deep, specialized benchmarks for core STEM disciplines. The \textbf{MATH} dataset \citep{hendrycksmath2021} was a landmark, providing challenging competition math problems with step-by-step solutions that have been pivotal for research \citep{wang2024mathvision}. To escalate the difficulty, \textbf{OlympiadBench} \citep{he-etal-2024-olympiadbench} incorporates problems from international Olympiads and the most difficult Gaokao questions. The extremely poor performance of models like GPT-4V (17.97\%) on this benchmark highlights its immense challenge and reveals critical failure modes in SOTA models \citep{zheng2024olympicarena}. \name draws inspiration from the focus on deep, complex reasoning inherent in these specialized benchmarks.

\subsection{Synthesis and Positioning of \name}

\name synthesizes the strengths of these three distinct trends. We adopt the methodological rigor and originality-first principles of frontier-difficulty benchmarks like GPQA. We draw on the focus on deep, complex reasoning from specialized STEM benchmarks like OlympiadBench. Our core contribution is to apply these principles to a novel and strategically important domain: a broad but coherent suite of AI for Science subjects. In doing so, \name fills a critical gap, providing a tool to measure and drive progress on the integrated reasoning skills vital for the next generation of scientific discovery \citep{luo2025llm4sr, zheng2025automation}. The table below provides a comparative summary.

\begin{table}[h]
\caption{High-Level Comparison of Benchmark Goals and Scope. We summarize prominent high-difficulty reasoning benchmarks, comparing their primary scientific goals and the scope of subjects they cover.}
\label{tab:benchmark_high_level}
\centering
\small 
\begin{tabularx}{\textwidth}{@{} l X X @{}}
\toprule
\textbf{Benchmark} & \textbf{Primary Goal} & \textbf{Subject Scope} \\
\midrule
MMLU & To measure multitask knowledge acquired during pretraining via zero- and few-shot evaluation across a wide range of subjects. & Covers 57 diverse subjects across STEM, humanities, and social sciences, from elementary to professional level. \\
\midrule
GPQA & To support scalable oversight research with ``Google-proof'' questions that are difficult for non-experts but verifiable by experts. & 448 graduate-level multiple-choice questions in Biology, Physics, and Chemistry. \\
\midrule
SuperGPQA & To scale GPQA-style evaluation to underrepresented and specialized ``long-tail'' academic disciplines. & Over 26,000 questions covering 285 graduate-level subjects, expanding well beyond traditional STEM. \\
\midrule
OlympiadBench & To test complex, multi-step reasoning on problems requiring creative and rigorous solutions, mirroring top-tier human competition. & Olympiad-level problems in Mathematics and Physics sourced from international and national competitions (e.g., IMO, IPhO). \\
\midrule
HLE & To measure knowledge at the frontier of human academic inquiry, addressing the performance saturation of earlier benchmarks like MMLU. & Highly challenging questions across Math, Science, Humanities, and CS, including multimodal and short-answer formats. \\
\midrule
\textbf{\name (Ours)} & \textbf{To measure frontier AI for Science (AI4S) reasoning, focusing on tasks central to the scientific discovery process with a rigorous ``human in loop'' pipeline.} & \textbf{Covers 7 core AI4S subjects and 47 subfields, with natural used questions for science research.} \\
\bottomrule
\end{tabularx}
\end{table}

\subsection{Additional Discussion of LLM-as-Judge}
Evaluating LLMs is now a central research focus, given their expanding deployment across applications ranging from natural-language processing to decision making. 
Conventional metrics often overlook the semantic and contextual subtleties of open-ended LLM outputs. Human assessment, while more reliable, is labor-intensive, costly, and hard to scale. 
The ``LLM-as-a-Judge'' paradigm has been proposed to overcome these limitations: an advanced LLM appraises the outputs of another model, yielding a scalable and economical proxy for human judgment. 
Foundational studies~\citep{ZhengC00WZL0LXZ23,abs-2310-02174} delineate both the promise and the limitations of this strategy; recent surveys~\citep{ChangWWWYZCYWWYZCYYX24,abs-2411-15594,abs-2412-05579} chart future directions. 
Continued progress will depend on mitigating bias, inconsistency, and prompt sensitivity to unlock the full potential of LLM-as-a-Judge systems.

\section{\name Construction Pipeline}

\begin{figure*}[t]
    \centering
    \includegraphics[width=\textwidth]{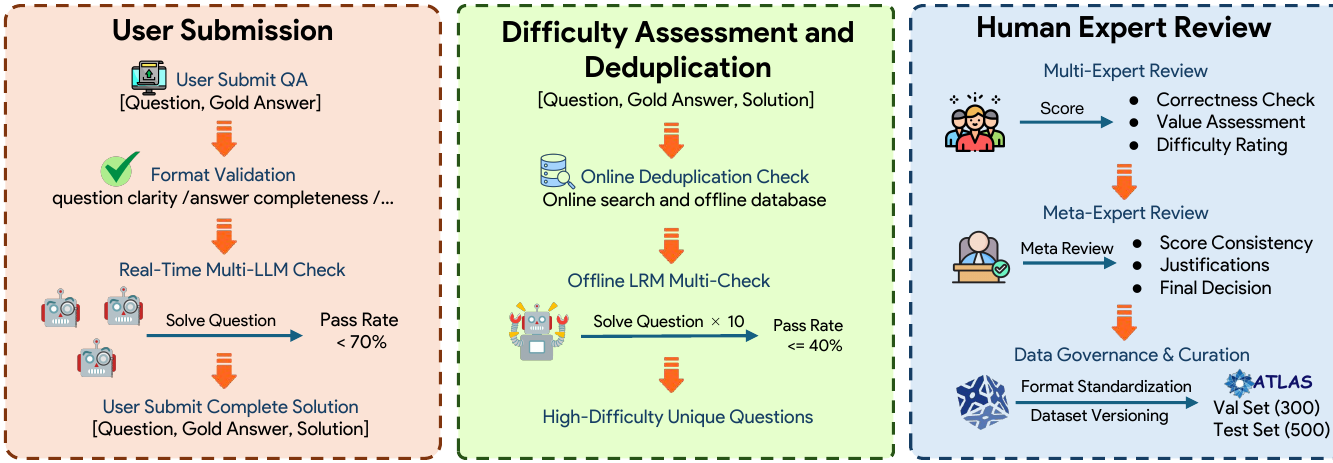}
    \caption{Overview of \name construction pipeline.}
    \label{fig:workflow}
\end{figure*}


In the current AI era, we recognize that the value of an evaluation benchmark depends not only on the questions it contains but, more importantly, on the methodological rigor in their creation. 
The methodology for constructing evaluation benchmarks has evolved from the straightforward data collection seen in MMLU to the complex, multi-stage, human-machine collaborative processes used by projects like GPQA and HLE, which are critical to ensuring their effectiveness and credibility~\citep{rein2023gpqa,du2025supergpqa}. 
In line with these advancements, we have designed and implemented a rigorous multi-stage process to systematically ensure the high quality, difficulty, and originality of the problems.

\subsection{Question Design for Real-World Fidelity}
Our question design prioritizes realism over evaluation convenience and adheres to the following principles:

\begin{itemize}[leftmargin=*]
    \item \textbf{Question Types. } We focus on short-answer and fill-in-the-blank formats. Over 50\% of the questions are \textbf{compound questions} with multiple sub-parts. 
This structure tests a model's ability to manage complex instructions, maintain long-range context, and perform multi-step reasoning, exposing weaknesses that single questions cannot.

    \item \textbf{Answer Complexity. } Answers are designed to be complex entities, such as a full \LaTeX{} equation ($\int_{0}^{\infty} e^{-x^2} dx = \frac{\sqrt{\pi}}{2}$), a list of chemical products, or a short, high-difficulty but simplified proof.

    \item \textbf{Bilingualism. } All questions in \name are available in both English and Chinese to support the global research community, a practice shared by other frontier benchmarks like OlympiadBench.
\end{itemize}

\subsection{Core Design Principles}
Our construction process is based on the following four core principles:

\begin{itemize}[leftmargin=*]
    \item \textbf{Frontier Difficulty and Originality.} To combat benchmark saturation and data contamination \citep{yue2025hle, rein2023gpqa}, all problems are newly-authored or substantially re-engineered by domain experts. Questions are targeted at a graduate-level or higher difficulty, explicitly testing complex, multi-step reasoning rather than information retrieval.

    \item \textbf{Hybrid Human-AI Quality and Difficulty Calibration.} We employ a rigorous, multi-stage validation pipeline. This includes a two-tiered, anonymous expert review process for scientific accuracy and clarity, inspired by methodologies from GPQA \citep{rein2023gpqa}. Crucially, this is augmented by an adversarial filtering stage where only problems that state-of-the-art models (e.g., DeepSeek-R1) fail to solve with high frequency (e.g., $\le$40\% accuracy) are retained, ensuring the benchmark remains at the frontier of AI capabilities.

    \item \textbf{Objective and Complex Answer Formulation.} To mirror real-world scientific outputs and prevent guessing, we eschew simple multiple-choice or short-string answers. Every problem features a single, objectively verifiable answer, often expressed in complex formats such as \LaTeX{} equations, chemical formulas, or structured multi-part responses, demanding generative reasoning rather than simple recognition.
\end{itemize}

\subsection{Data Generation and Quality Assurance Workflow}
Figure \ref{fig:workflow} details our data construction and quality assurance workflow. This process combines the deep domain knowledge of human experts with the computational power of large models, forming a powerful ``dual-filter” system to ensure that every problem admitted to the final database possesses both high quality and high difficulty.
The workflow includes the following key stages:

\begin{figure*}[t]
    \centering
    \includegraphics[width=\textwidth]{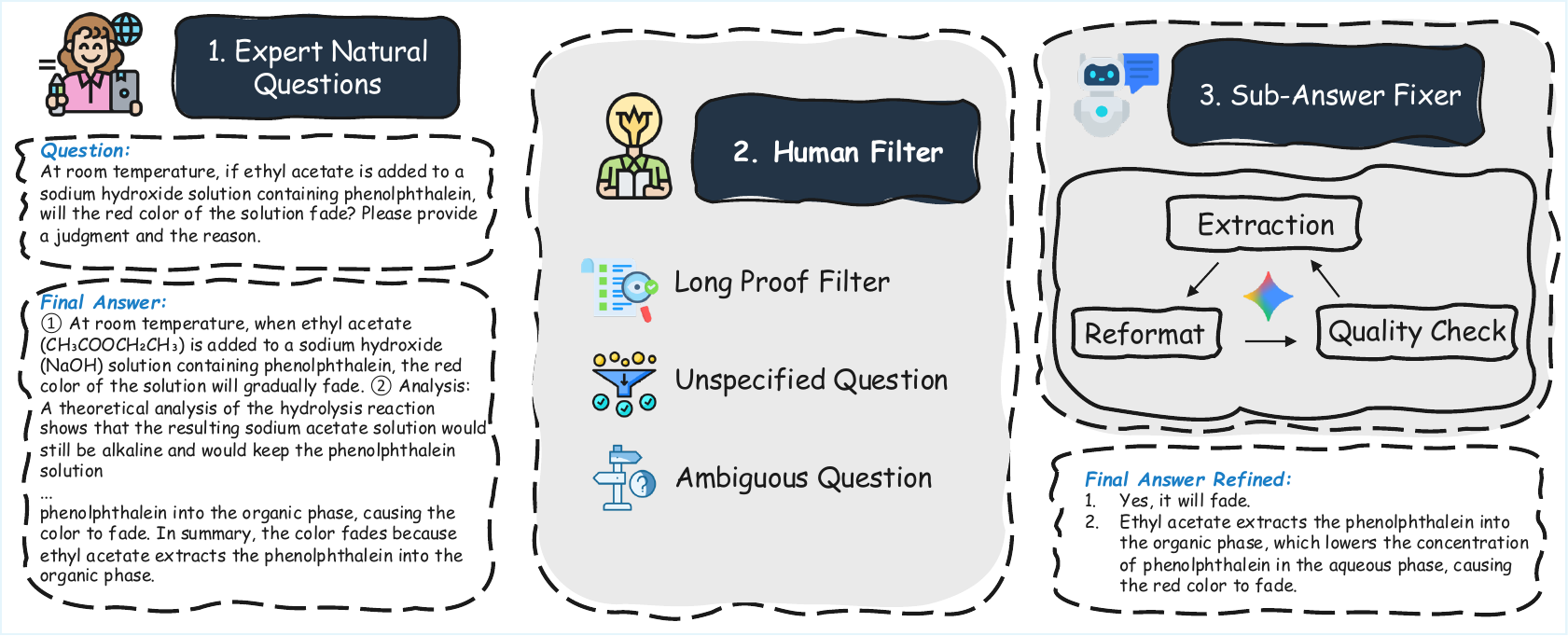}
    \caption{Overview of how \name refine the final answer for natural questions. }
    \label{fig:answer_refine_pipe}
\end{figure*}

\begin{itemize}[leftmargin=*]
    \item \textbf{Stage 1: Expert-Sourced Problem Generation and Pre-screening. } Our process begins with Ph.D.-level experts from more than 25 different institutions crafting original problems that demand multi-step, and often cross-disciplinary, reasoning. Each submission includes a canonical solution and detailed steps. These problems then undergo an automated pre-screening pipeline, which normalizes formatting and performs a similarity check against a vast offline database of existing problems to filter out duplicates and ensure novelty.

    \item \textbf{Stage 2: Adversarial Filtering and Iterative Refinement. } To ensure the questions in our benchmark are both novel and sufficiently challenging, we implement a rigorous pipeline for adversarial filtering and iterative refinement. This process consists of two main phases: originality verification and difficulty calibration. First, to mitigate data contamination, each submission undergoes an originality assessment. We employ a Retrieval-Augmented Generation~(RAG) system to screen submissions against a comprehensive corpus of web content, academic papers, and existing benchmarks. This system first retrieves the top-$K$ most semantically similar entries. Subsequently, an LLM is used to evaluate these retrieved items and assign both redundancy and originality scores to the submission. Only submissions that meet a high originality threshold advance to the next stage.  Following the originality check, problems are evaluated for difficulty by an ensemble of state-of-the-art LRMs. We apply a stringent adversarial criterion: a problem is accepted only if these LRMs achieve a solution accuracy of 40\% or less over ten attempts. This strict standard ensures that the final problems robustly challenge current AI capabilities. Problems that do not meet this difficulty threshold are returned to the human experts. They can then choose to discard the problem or iteratively refine it to increase its complexity before resubmission, creating a closed-loop quality enhancement process.

    \item \textbf{Stage 3: Multi-Layered Human Validation and Final Ingestion. } Problems that pass the adversarial filtering undergo a rigorous, multi-stage manual quality inspection. Each problem is sent to three anonymous peer reviewers in the same domain for a double-blind evaluation of its correctness, clarity, and difficulty. Discrepancies in reviews are resolved by a senior meta reviewer who makes a final determination. Finally, before being admitted to the benchmark, a last check is performed against online search engines to confirm the problem has not been publicly disclosed. Only problems that clear every stage of this comprehensive validation process are accepted into the final database. Detailed information about human review stage can refer to \Cref{expert_review}.

    \item \textbf{Stage 4: Final Answer Refinement and Verification. } Following the rigorous validation of the problems, as shown in \Cref{fig:answer_refine_pipe}, a final stage is dedicated to refining the expert-provided answers to ensure maximum clarity, correctness, and pedagogical value. This process, illustrated in the provided figure, transforms the initial expert solutions into a canonical format suitable for the benchmark. The refinement pipeline consists of three key steps: First, a LLM agent performs \textbf{Extraction}, decomposing the initial, often verbose, answer into its fundamental components, such as the direct judgment (e.g., ``the color will fade'') and the scientific reasoning. Next, the extracted components undergo a semi-automated \textbf{Quality Check} and \textbf{Reformatting} process. During this step, the agent verifies the factual and scientific accuracy of the underlying reasoning—for instance, correcting an initial hypothesis of hydrolysis to the correct mechanism of solvent extraction as shown in the example. Concurrently, the answer is restructured into a clear, step-by-step format, eliminating ambiguities and extraneous details. This ensures that the final refined answer is not only correct but also presented in a structured and easily digestible manner, thereby enhancing its utility for precise model evaluation.
\end{itemize}

This unique dual-filter system, which combines adversarial LLM filtering with multi-stage manual review, provides a robust operational definition for ``high-quality difficulty''.
The LLM filtering ensures that problems are challenging for machines, building on the successful experiences of projects like HLE~\citep{yue2025hle}. 
The multi-stage manual review ensures that this difficulty stems from the problem's scientific depth and complexity, rather than from flaws, ambiguities, or reliance on obscure knowledge, aligning with GPQA's focus on human expert performance~\citep{rein2023gpqa}. 
This hybrid methodology is essential for ensuring the long-term validity and credibility of \name.

\section{Dataset Analysis: The \name Corpus}
The first phase of the \name project has resulted in the \name corpus, which contains approximately 800 high-quality scientific reasoning problems selected through our rigorous process. 
This section provides a detailed analysis of this dataset from both quantitative and qualitative perspectives.

\subsection{Quantitative Overview}
\name covers seven core disciplines of AI for Science. To provide a comprehensive evaluation, we have established several sub-fields under each discipline and ensured a balanced distribution of text-only and multimodal problems. The table below shows the detailed statistical distribution of the dataset.


\begin{table}[h]
\caption{Statistical Details of the Benchmarks}
\label{tab:benchmark_details}
\centering
\begin{tabular}{llr}
\toprule
\textbf{Category} & \textbf{Sub-category} & \textbf{Count / Percentage} \\
\midrule
\multicolumn{2}{l}{\textbf{\name (Breakdown by Subject)}} & \textbf{798} \\
& Physics & 175 \\
& Materials Sci. & 140 \\
& Chemistry & 117 \\
& Earth Sci. & 109 \\
& Biology & 102 \\
& Mathematics & 94 \\
& Computer Science & 61 \\
\midrule
\multicolumn{2}{l}{\textbf{\name (Breakdown by Question Type)}} & \textbf{100\%} \\
& Calculation \& Derivation & 71.4\% \\
& Selection \& Judgment & 12.2\% \\
& Explanatory \& Descriptive & 10.2\% \\
& Structured \& Composite & 6.1\% \\
\bottomrule
\end{tabular}
\end{table}
\subsection{Qualitative Examples}
To give readers a more intuitive feel for the characteristics of the problems in \name, we present a few representative examples from different subjects shown in Question \ref{qn:math_example} and Question \ref{qn:biolo_example}.

\begin{question}{Mathematics Example}{math_example}
\begin{itemize}[noitemsep,leftmargin=*]
    \item \textbf{Sub-field:} 
    Algebra and Geometry
    \item \textbf{Problem:}
    Let $p$ be an odd prime, and let $m \geq 0$ and $N \geq 1$ be integers. Let $\Lambda$ be a free $\mathbb{Z}/p^N\mathbb{Z}$-module of rank $2m + 1$, and let \[ (,): \Lambda \times \Lambda \to \mathbb{Z}/p^N\mathbb{Z} \] be a perfect symmetric $\mathbb{Z}/p^N\mathbb{Z}$-bilinear form. Here, 'perfect' means that the induced map \[ \Lambda \to \text{Hom}_{\mathbb{Z}/p^N\mathbb{Z}}(\Lambda, \mathbb{Z}/p^N\mathbb{Z}), \quad x \mapsto (x, \cdot) \] is an isomorphism. Find the number of elements in the set \[ \{ x \in \Lambda \mid (x, x) = 0 \} \] as a function of $p, m, N$.
    
    \item \textbf{Solution:} 
    For each integer $0 \leq n \leq N$, let $\Lambda(n) := \{ x \in \Lambda \mid (x, x) \in p^n \mathbb{Z}/p^N \mathbb{Z} \}$. Let $C(n) := |\Lambda(n)|$. We want to compute $C(N)$. It is trivial that $C(0) = |\Lambda| = p^{(2m+1)N}$
    ...
    We can establish two claims: 1. For $n \geq 2$, the multiplication-by-$p$ map $\Lambda(n-2)/p^{n-1}\Lambda \to 
    \Lambda(n)''/p^n\Lambda$ is a bijection. 2. For $n \geq 2$, the map $\Lambda(n)'/p^n\Lambda \to \Lambda(n-1)'/p^{n-1}\Lambda$ is $p^{2m}$-to-1.
    These claims lead to the recurrence relation:
    $ C(n) = C(n)' + C(n)'' = p^{-(2m+1)} C(n-2) + p^{(2m+1)(N-1)-(n-1)} (p^{2m} - 1). $
    Solving this recurrence yields the final result for $C(N)$.

    \item \textbf{Refined Final Answer:} 
    $p^{(2m+1)r + 2m(N-2r)} + \frac{p^{(2m+1)r} - 1}{p^{(2m+1)} - 1} p^{(2m+1)r-1 + 2m(N-2r)} (p^{2m} - 1)$, where $r := \lfloor N/2 \rfloor$.

    \item \textbf{Source Organization:} Fudan University
\end{itemize}
\end{question}

\begin{question}{Biology Example}{biolo_example}
    \begin{itemize}[noitemsep,leftmargin=*]
        \item \textbf{Sub-field:} 
        Immunology
        \item \textbf{Problem:}
        Background: In the innate immune system, RIG-I-like receptor (RLR) family proteins recognize viral RNA in the cytoplasm, triggering the downstream mitochondrial antiviral signaling protein (MAVS). MAVS acts as a signaling adapter, recruiting multiple proteins to form the MAVS signalosome, which activates transcription factors IRF3 and NF-$\kappa$B, inducing the expression of type I and type III interferons (IFNs) and other antiviral genes.        
        \begin{enumerate}[label=\arabic*., topsep=0pt, itemsep=0pt, partopsep=0pt, parsep=0pt]
            \item What is the core RNA-binding region of MAVS?
            \item In the interaction mechanism between the key adapter protein MAVS (mitochondrial antiviral signaling protein) and cellular RNA in innate immunity, what part of the cellular mRNA does MAVS directly bind to via its central disordered domain to regulate downstream antiviral signal transduction of RIG-I-like receptors (RLRs)?
            \item Treatment with RNase disrupts the stability of what complex, and reduces what property of transcription factors like IRF3 and NF-$\kappa$B p65, indicating that cellular RNA is crucial for the activation and formation of the MAVS signalosome?
        \end{enumerate}
        
        \item \textbf{Refined Final Answer:} 
          \begin{enumerate}[label=\arabic*.]
            \item Central disordered region
            \item 3'UTR (3' Untranslated Region)
            \item MAVS signalosome complex; phosphorylation level
          \end{enumerate}
    \item \textbf{Source Organization:} Shanghai Jiao Tong University School of Medicine
  \end{itemize}
  \end{question}

\name distinguishes itself by focusing on problems that demand a synthesis of expert-level domain knowledge and complex, multi-step reasoning chains. The generative, short-answer format fundamentally prevents ``guessing” and forces models to construct answers from first principles. For instance, the mathematics problem shown requires not just recalling definitions from abstract algebra but performing a multi-step, non-trivial derivation involving recurrence relations over a finite ring $\mathbb{Z}/p^N\mathbb{Z}$. This probes a model's ability to manipulate abstract symbolic structures.

Furthermore, the problems often necessitate causal and mechanistic reasoning, as seen in the biology example. To answer correctly, a model must navigate the complex cascade of the MAVS signaling pathway, identifying specific molecular components (central disordered region), their binding targets (3'UTR), and the functional consequences of their interactions (phosphorylation). This requires integrating disparate facts into a coherent causal model, a hallmark of true scientific understanding. Many problems in the corpus are not self-contained; they implicitly assume a knowledge base equivalent to that of an advanced undergraduate or graduate student in the field. This knowledge-intensive nature, combined with the demand for rigorous, generative reasoning, establishes \name as a challenging and realistic benchmark for evaluating the capabilities of next-generation AI models in the scientific domain.

\subsection{Language and Structural Characteristics}
The problems in \name are also challenging in terms of language and structure. The average length of a problem statement is about 65 words, but some problems describing complex scenarios can exceed 200 words. 
The answer format is short answer or fill-in-the-blank, requiring the model to generate precise text, numerical values, or mathematical expressions in \LaTeX~ format. 
The extensive use of LaTeX (especially in physics and mathematics problems) places higher demands on the model's ability to generate and understand symbols. Compared to multiple-choice questions, this generative evaluation method can more effectively prevent models from scoring by guessing, thus more accurately reflecting their reasoning and expression abilities.

\section{Evaluation and Performance}
In this section, we conduct an extensive evaluation of \name. We firstly establish a standardized evaluation framework to assess the performance of LLMs. 
Subsequently, we evaluate several leading LLMs and provide a comprehensive analysis.

\subsection{Setup}

\paragraph{LLMs. }
We encompass a representative series of frontier large reasoning models for evaluation, incorporating both closed-source proprietary models and prominent open-source models. 
The examined closed LRMs include: OpenAI GPT-5~\citep{openai2025gpt5}, OpenAI o3~\citep{openai2024o3}, OpenAI o4-mini, Gemini-2.5-Pro~\citep{abs-2507-06261}, Grok-4~\citep{xai2025grok4}, and Doubao-Seed-1.6-thinking~\citep{bytedance2025seed1_6}, as well as open-source LRMs such as DeepSeek-V3.1~\citep{abs-2412-19437}, GPT-OSS-120B~\citep{abs-2508-10925}, DeepSeek-R1-0528~\citep{abs-2501-12948}, Qwen3-235B-A22B~\citep{abs-2505-09388}, Qwen3-235B-A22B-2507~\citep{abs-2505-09388}, and GLM-4.5~\citep{abs-2508-06471}.

\paragraph{Judge as Reasoning. }
As previously noted, the answers in \name comprise multiple responses, along with complex natural language and symbolic descriptions, which complicate the assessment of the alignment between model predictions and true answers using rule-based heuristic methods~\citep{abs-2412-05579,ZhengC00WZL0LXZ23,liu2025compassverifier}. 
To address this challenge, we regard the evaluation of \name as a complex reasoning task, employing prominent large reasoning models to evaluate the model prediction results. 
In this paper, we utilize two models, OpenAI o4-mini~\citep{openai2024o3} and GPT-OSS-120B~\citep{abs-2508-10925}, as Judge models.

\paragraph{Metrics. }
Referencing typical reasoning tasks such as code~\citep{abs-2107-03374} and mathematics~\citep{hendrycksmath2021,openai2024o3,abs-2501-12948}, we report the average accuracy across multiple inferences and G-Pass@$k$~\citep{abs-2412-13147} to assess the stability of LLMs' performance.

\paragraph{Implementation Details. }
For the closed-source LLMs~(i.e., LRMs), we utilize the official API to obtain predictions, while for the open-source LLMs, we deploy them using serving frameworks like SGLang~\citep{ZhengYXS0YCKSGB24} and vLLM~\citep{KwonLZ0ZY0ZS23} for inference. 
We set the maximum number of generation tokens for each LLM to 32,768, and the sampling temperature is established at 0.6.
For each question, we generate 4 predictions. 
The complete experiment roughly consumed all the API quota worth \$3,000, as well as hundreds of GPU Hours.

\subsection{Evaluation Workflow}
\begin{figure*}[t]
    \centering
    \includegraphics[width=1.\textwidth]{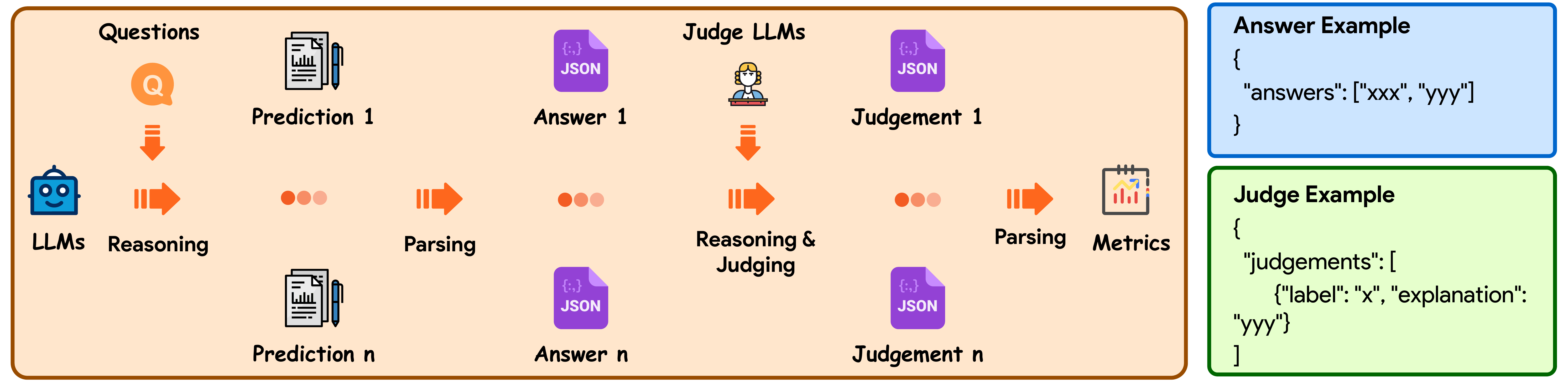}
    \caption{Overview of the evaluation workflow. During the evaluation process, the LLM is prompted to provide formatted predictions, from which the answers are extracted and input into the Judge LLMs for the computation of evaluation metrics.}
    \label{fig:eval_workflow}
\end{figure*}

Considering the complexity of evaluation of \name, it is difficult to evaluate the model's performance through simple and conventional evaluation processes.
We propose a comprehensive and user-friendly evaluation framework for assessing \name based on the OpenCompass~\citep{contributors2023opencompass} repository.
As illustrated in \cref{fig:eval_workflow}, the evaluation workflow consists of the following steps: 
1) Prediction Generation; 
2) Answer Parsing; 
3) Judgment Generation; 
4) Judgment Parsing.

\paragraph{Step 1: Prediction Generation. }
We provide each LLM~(i.e., LRM) a detailed instruction to generate predictions as shown in Prompt \ref{pp:prompt_inference}.
The LLM is prompted to solve the given question step by step and must put their final answers in the JSON format.
The advantages of this approach are as follows, which facilitates the extraction of answers, particularly for questions with multiple sub-questions.

\paragraph{Step 2: Answer Parsing. }
After obtaining the predictions from the LLM, we parse the JSON-formatted answers to extract the final answers.

\paragraph{Step 3: Judgment Generation. }
During this step, we input the original question, the parsed answer, and the ground truth into the Judge LLM. 
The Judge LLM generates assessments based on the instructions provided in Prompt \ref{pp:prompt_judgment}. 
Our instructions direct the LLM to evaluate the correctness of each sub-answer while considering reasonable error, providing its judgments for each sub-answer in JSON format.

\paragraph{Step 4: Judgment Parsing. }
Similar to the parsing of the answers, we parse the JSON-formatted judgments and calculate the evaluation metrics.

Our evaluation workflow offers insights into the assessment of complex problems. 
It not only provides judgment results for the overall outcome but also delivers fine-grained judgment results, which are beneficial for applications in Reinforcement Learning with Verifiable Rewards.

\subsection{Quantitative Results}
In this section, we present the quantitative results and analysis of the evaluation conducted on the validation set of \name. For more detailed results on the test set of \name, please refer to \Cref{app:test_performance}.

\begin{table}[t]
    \centering
    \caption{The performance of various LLMs on the validation set of \name, as judged by GPT-OSS-120B, is sorted by average accuracy. Each LLM is prompted to generate four predictions, and we report the average accuracy as well as the mG-Pass@$\{2,4\}$ scores. A high mG-Pass score indicates a high level of stability across multiple predictions.} \label{tab:model_performance_judge_oss}
    \resizebox{\textwidth}{!}{
        \begin{tabular}{lcrrr}
            \toprule
            \textbf{Model} & \textbf{\#Tokens} & \textbf{Accuracy (\%)} $\uparrow$ & \textbf{mG-Pass@2 (\%)} $\uparrow$ & \textbf{mG-Pass@4 (\%)} $\uparrow$ \\
            \midrule
            OpenAI GPT-5-High & 32k & 42.9 & 34.7 & 32.1 \\
            Gemini-2.5-Pro & 32k & 35.3 & 25.3 & 23.4 \\
            Grok-4 & 32k & 34.1 & 25.8 & 24.1 \\
            OpenAI o3-High & 32k & 33.8 & 24.0 & 22.3 \\
            DeepSeek-R1-0528 & 32k & 26.4 & 16.1 & 14.1 \\
            Qwen3-235B-A22B-2507 & 32k & 26.1 & 18.4 & 16.9 \\
            Doubao-Seed-1.6-thinking & 32k & 26.1 & 18.1 & 16.8 \\
            DeepSeek-V3.1 & 32k & 25.3 & 16.3 & 15.0 \\
            OpenAI o4-mini & 32k & 22.4 & 15.1 & 13.5 \\
            GPT-OSS-120B-High & 32k & 21.7 & 14.1 & 12.8 \\
            \bottomrule
        \end{tabular}
    }
\end{table}

\paragraph{Overall Performance. }
The evaluation performance show in \Cref{tab:model_performance_judge_oss} on the validation set of \name reveals that: 
1) OpenAI GPT-5-High stands out as the top-performing model, achieving the highest accuracy (42.9\%) and exhibiting strong prediction stability, with mG-Pass@2 at 34.7\% and mG-Pass@4 at 32.1\%;
2) The mG-Pass scores corroborate the accuracy results, indicating that models with higher accuracy typically exhibit greater stability across multiple predictions; 
3) A notable performance disparity exists between the top-tier models (OpenAI GPT-5, OpenAI o3, Gemini-2.5-Pro, Grok-4).
Specifically, OpenAI o3-High ranks second with an accuracy of 35.3\%, maintaining stability with mG-Pass@2 at 25.3\% and mG-Pass@4 at 23.4\%. 
Gemini-2.5-Pro closely follows, recording an accuracy of 34.1\% and mG-Pass scores of 25.8\% (@2) and 24.1\% (@4), indicating competitive stability. 
The remaining models, such as DeepSeek-R1-0528 (26.4\%), DeepSeek-V3.1 (25.3\%), Qwen3-235B-A22B-2507 (26.1\%), Doubao-Seed-1.6-thinking (26.1\%), OpenAI o4-mini (22.4\%), and GPT-OSS-120B-High (21.7\%) exhibit progressively lower accuracy and stability. 
Notably, some open-source LLMs like DeepSeek-R1-0528 and Qwen3-235B-A22B-2507 still deliver competitive results compared to other proprietary systems in the lower tier.

\begin{figure}[t!]
    \centering
    \includegraphics[width=\linewidth]{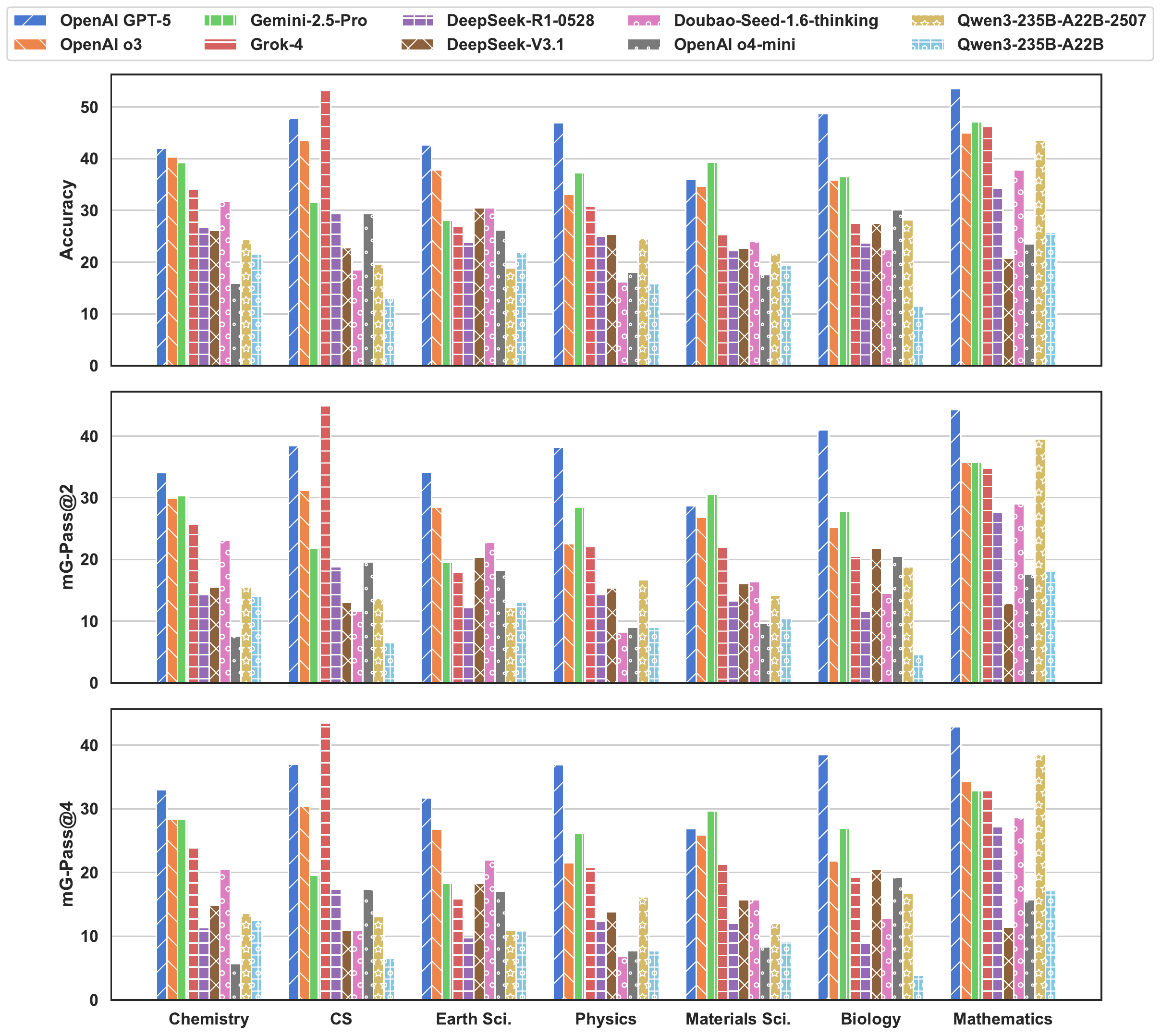}
    \caption{The performance of different LLMs across different subjects of \name's validation set.} \label{fig:subject_performance}
\end{figure}

\paragraph{Subject Performance. }
\Cref{fig:subject_performance} demonstrates the performance of different LLMs across different subjects of \name's validation set. 
Across all subjects and metrics, OpenAI GPT-5 consistently leads, achieving the highest accuracy and mG-Pass scores by a clear margin. 
Gemini-2.5-Pro also produces competitive results, particularly in Physics, Chemistry, and Biology. 
Grok-4 shows notable strength in CS, achieving the highest performance in this domain. 
In contrast, Qwen3-235B-A22B-2507 and Qwen3-235B-A22B generally exhibit the lowest scores across most subjects and metrics, indicating weaker performance in this \name evaluation. 
OpenAI o3 and DeepSeek-V3.1 display moderate performance, while Doubao-Seed-1.6-thinking yields mixed results, performing relatively well in some areas but lagging in others.
For Specific Subjects:
\begin{itemize}[leftmargin=*]
    \item \textbf{Chemistry}: OpenAI GPT-5 clearly dominates, followed by Gemini-2.5-Pro, while Grok-4 and Doubao-Seed-1.6-thinking perform moderately well.
    \item \textbf{CS}: Grok-4 significantly outperforms all other models, achieving the highest accuracy and mG-Pass scores, with GPT-5 and o3 trailing behind.
    \item \textbf{Earth Sci.}: OpenAI GPT-5 leads with the highest accuracy and stability, while o3 and Gemini-2.5-Pro achieve moderate performance.
    \item \textbf{Physics}: GPT-5 achieves the strongest results, with Gemini-2.5-Pro and o3 also showing competitive performance.
    \item \textbf{Materials Sci.}: GPT-5 dominates this domain, while Gemini-2.5-Pro and o3 form the second tier of performance.
    \item \textbf{Biology}: GPT-5 again leads by a wide margin, with Gemini-2.5-Pro and Grok-4 performing moderately well.
    \item \textbf{Mathematics}: GPT-5 achieves the highest performance, followed by Qwen3-235B-A22B-2507, which shows competitive results, while Gemini-2.5-Pro also performs strongly. 
\end{itemize}
Finally, we observe that the mG-Pass@$\{2,4\}$ scores generally align with the trend of accuracy, indicating that models with higher accuracy also demonstrate greater inference stability. 
GPT-5's consistently high mG-Pass@$\{2,4\}$ scores across all domains underscore its leading performance, while Grok-4's dominance in CS highlights its particular strength in this field. 
Conversely, the lower mG-Pass@$\{2,4\}$ scores for models such as Qwen3-235B-A22B indicate not only reduced accuracy but also less stable or more variable predictions.

\subsection{Further Analysis}

\begin{table}[t]
    \centering
    \caption{Answer extraction error rate of different LLMs on \name.} \label{tab:extraction_error}
    \begin{tabular}{lcc}
        \toprule
        \textbf{Model} & \textbf{Truncation Rate(\%)} & \textbf{JSON Parse Error Rate (\%)} $\downarrow$ \\
        \midrule
        OpenAI o4-mini & 0.00 & 0.00\\
        OpenAI o3 & 1.58 & 0.00 \\
        DeepSeek-R1-0528 & 2.16 & 0.00 \\
        Gemini-2.5-Pro & 3.49 & 0.00 \\
        Doubao-Seed-1.6-thinking & 8.22 & 0.08 \\
        Grok-4 & 10.38 & 0.00 \\
        \bottomrule
    \end{tabular}
\end{table}
\begin{table}[t]
    \centering
    \caption{The performance of selected LLMs on the validation set of \name under 64k and 32k coutput budget, as judged by GPT-OSS-120B, is sorted by average accuracy. } \label{tab:model_performance_64k}
    \resizebox{\textwidth}{!}{
        \begin{tabular}{lcrrr}
            \toprule
            \textbf{Model} & \textbf{\#Tokens} & \textbf{Accuracy (\%)} $\uparrow$ & \textbf{mG-Pass@2 (\%)} $\uparrow$ & \textbf{mG-Pass@4 (\%)} $\uparrow$ \\
            \midrule
            \multirow{2}{*}{OpenAI GPT-5-High} & 64k & 43.7 & 35.2 & 33.6 \\
            & 32k & 42.9 & 34.7 & 32.1 \\
            \midrule
            \multirow{2}{*}{OpenAI o3-High} & 64k & 36.6 & 26.6 & 25.3 \\
            & 32k & 33.8 & 24.0 & 22.3 \\
            \midrule
            \multirow{2}{*}{Gemini-2.5-Pro} & 64k & 36.6 & 27.7 & 26.1 \\
            & 32k & 35.3 & 25.3 & 23.4 \\
            \midrule
            \multirow{2}{*}{Qwen3-235B-A22B-2507} & 64k & 27.6 & 20.5 & 19.3 \\
            & 32k & 26.1 & 18.4 & 16.9 \\
            \midrule
            \multirow{2}{*}{DeepSeek-V3.1} & 64k & 26.4 & 17.3 & 15.5 \\
            & 32k & 25.3 & 16.3 & 15.0 \\
            \bottomrule
        \end{tabular}
    }
\end{table}

\paragraph{Output Budget and Answer Extraction. }
During the evaluation process, the inability to extract answers can detrimentally affect the model's performance. In our evaluation workflow, the main extraction errors originate from prediction truncation and JSON parsing errors. 
Consequently, we have documented the rates of answer extraction errors across various LLMs, as illustrated in \Cref{tab:extraction_error}. 
A notable observation from the table is that almost models achieved a 0.00\% JSON Parse Error rate. 
This result is excellent, indicating that once an answer is generated, the JSON output structure remains consistently valid across all evaluated LLMs. 
This result implies robust parsing capabilities or careful adherence to JSON formatting of current salient LLMs, which is crucial for automated judgment and verification in data synthesis and reinforcement learning. 
Conversely, the Truncation Rate reveals significant variations in performance. 
OpenAI o4-mini is exceptional, exhibiting a 0.00\% Truncation Rate, indicating it never truncates its answers, thereby ensuring complete responses—an essential characteristic for applications requiring comprehensive information. 
OpenAI o3 also demonstrates very low truncation rates at 1.58\%, indicating that it rarely truncates its answers. 
While not perfect, these rates are commendable, suggesting that the majority of its answers are complete. 
DeepSeek-R1-0528 and Gemini-2.5-Pro display moderate truncation rates of 2.16\% and 3.49\%, respectively. 
The models with the highest truncation rates are Doubao-Seed-1.6-thinking at 8.22\% and Grok-4 at 10.38\%. 
These statistics are concerning, as they imply that over 8\% and 10\% of their generated answers are truncated, respectively. 
This suggests that the models may produce excessively lengthy chains of thought and necessitate improvements in their efficiency.
To illustrate the impact of output budget, we conducted experiments with a token budget of 64k compared to 32k, as detailed in \Cref{tab:model_performance_judge_oss}. 
As presented in \Cref{tab:model_performance_64k}, while most LLMs exhibit improved performance with increased limits from 32k to 64k output tokens, this extension in output length incurs significant inference overhead, particularly given the parameter size of contemporary LLMs. 
This underscores the importance of enhancing the inference efficiency of LLMs.

\begin{table}[t]
    \centering
    \caption{Summary of the primary error categories of \name. We randomly sample 200 judge explanations of erroneous predictions to identify and summarize the most frequent error modes.} \label{tab:error_mode}
    \begin{tabular}{l>{\raggedright\arraybackslash}p{0.45\textwidth}cc}
        \toprule
        \textbf{Error Category} & \textbf{Description} & \textbf{Proportion (\%)} \\
        \midrule
        Numerical discrepancies & Numerical values differ from the standard beyond acceptable tolerance (e.g., $\pm 0.1$). & 27.0 \\
        Mathematical errors & Incorrect formulas, equations, or mathematical expressions (e.g., wrong coefficients, terms). & 16.5 \\
        Missing components & Omission of required elements (e.g., terms in equations, methods, or parts of multi-part answers). & 13.0 \\
        Structural mismatches & Answers differ in format, structure, or functional form from the standard answer. & 11.0 \\
        Incorrect methods & Use of wrong approaches, techniques, or assumptions to solve the problem. & 8.5 \\
        Contradictory conclusions & Answers directly oppose the standard answer's conclusion or assertion (e.g., ``Yes'' vs. ``No''). & 7.0 \\
        Unit errors & Missing, incorrect, or inconsistent units in the answer. & 5.0 \\
        Incorrect reasoning & Flawed logic or explanations that do not align with the standard answer's reasoning. & 4.5 \\
        Misinterpretation of concepts & Misunderstanding of key principles or concepts required for the answer. & 4.0 \\
        Incomplete answers & Partially correct answers lacking essential details or components. & 3.5 \\
        \bottomrule
    \end{tabular}
\end{table}

\paragraph{Error Category. }
To provide valuable insights into areas where improvement efforts would have the most impact, we analyze the errors in the prediction results, as summarized in \Cref{tab:error_mode}. 
The results reveal that numerical discrepancies are the most prevalent error category, accounting for 27.0\% of all errors. 
This finding suggests that precision in numerical outputs or calculations poses a significant challenge. 
Following numerical discrepancies, mathematical errors represent the second largest category at 16.5\%, indicating difficulties with the application of correct formulas, equations, or expressions. 
Missing components (13.0\%) and structural mismatches (11.0\%) are also significant error types, underscoring issues related to completeness and adherence to expected output formats. 
Additionally, incorrect methods and reasoning account for a notable proportion, suggesting substantial room for improvement in the current LLMs' professional knowledge within the scientific domain.

\begin{table}[t]
    \centering
    \caption{The performance comparison between judged by Qwen3-235B-A22B and GPT-OSS-120B is as follows: a "+" subscript indicates that the score judge by Qwen3-235B-A22B outperforms that by GPT-OSS-120B, while a "-" signifies the opposite.}
    \label{tab:model_performance_judge_qwen3}
    \begin{tabular}{lccc}
    \toprule
    \textbf{Model} & \textbf{Accuracy (\%)} $\uparrow$ & \textbf{mG-Pass@2 (\%)} $\uparrow$ & \textbf{mG-Pass@4 (\%)} $\uparrow$ \\
    \midrule
    OpenAI o3 & 30.3$_{-5.3}$ & 19.3$_{-6.4}$ & 17.3$_{-6.9}$ \\
    Gemini-2.5-Pro  & 31.3$_{-3.6}$ & 21.0$_{-3.4}$ & 19.3$_{-3.2}$ \\
    Grok-4 & 30.7$_{-2.2}$ & 21.8$_{-3.2}$ & 19.9$_{-3.6}$ \\
    DeepSeek-R1-0528  & 22.0$_{-3.8}$ & 12.9$_{-2.6}$ & 11.3$_{-2.1}$ \\
    \bottomrule
    \end{tabular}
\end{table}

\paragraph{Judge Model.}
Given that our evaluation is highly related to the judge model used, we analyze the performance of various advanced reasoning models used as judge models. 
As demonstrated in \Cref{tab:model_performance_judge_qwen3}, we compare the performance evaluations conducted by Qwen3-235B-A22B and GPT-OSS-120B, respectively. 
Across nearly all models and metrics, Qwen3-235B-A22B typically allocates lower scores than GPT-OSS-120B in the domains of Accuracy, mG-Pass@2, and mG-Pass@4. 
To investigate these differences, we conducted a case analysis comparing the judgments of GPT-OSS-120B and Qwen3-235B-A22B. 
As demonstrated in Case \ref{cs:case_cs}, which involves a computer science question in the validation set of \name, OpenAI o3 produced the prediction $t_n = 2n\ln n(1+o(1))$. 
In the context of algorithm complexity in computer science, $\log$ and $\ln$ are equivalent, verifying the correctness of OpenAI o3's prediction. 
However, Qwen3-235B-A22B fails to acknowledge this equivalence, leading to an incorrect judgment. 
Additionally, in Case \ref{cs:case_materials}, which shows a Materials Sci. question, OpenAI o3 accurately predicts the outcomes for two sub-questions, whereas Qwen3-235B-A22B erroneously assumed that the prediction ``yes,no'' pertained to a single question, resulting in an incorrect judgment.
Lastly, as illustrated in Case \ref{cs:case_physics}, involving a physics question, OpenAI o3 provided an answer of $1.6 \times 10^2 \text{N}$, exhibiting a relative error of 0.376\% compared to the standard answer, thus meeting the permissible error conditions specified in our judge prompt~(Prompt \ref{pp:prompt_judgment}). 
Consequently, it should be deemed correct. 
However, Qwen3-235B-A22B erroneously identified the absolute error as the relative error, resulting in an incorrect judgment.
The case analysis results indicate that reasoning models with advanced capabilities also exhibit superior judgment accuracy. 
Compared to GPT-OSS-120B, Qwen3-235B-A22B is more susceptible to errors in knowledge and semantic understanding. 
Nonetheless, assessing the effectiveness of judgment models requires further investigation, which falls outside the scope of this paper. 
We hope the introduction of \name will encourage the community to advance research on judgment models for questions that are difficult to verify.

\begin{case}{Case in CS subject}{case_cs}
    \subsection*{Question: }
    Given an array $A[1:n]$ of $n$ distinct numbers, recall the *quicksort* algorithm to sort them into an ascending order.

    \begin{itemize}[leftmargin=*]
        \item First we let $p\gets A[1]$ be the pivot.
        \item Then we use $p$ to rearrange $A[1:n]$ in the following way:
        \begin{itemize}[leftmargin=*]
            \item For some $k\in [n]$, we have $A[k]=p$;
            \item For all $i\in \{1,2,\dots,k-1\}$, we have $A[i]<p$;
            \item For all $i\in \{k+1,\dots,n\}$, we have $A[i]>p$.
            \item Then recursively sort $A[1:k-1]$ and $A[k+1;n]$.
        \end{itemize}
    \end{itemize}

    Assume our algorithm used in step 2 to compute $A[1:k-1]$ and $A[k+1:n]$ is *stable*. That is, if $A[i]$ and $A[j]$ are both smaller or larger than $p$ and $i<j$, then after the rearrangement, $A[i]$ still appear before $A[j]$.

    Clearly the performance of the algorithm depends on the input array $A$. In this problem, we analyze its *average complexity*. Suppose that the input array $A$ is chosen uniformly at random from the $n!$ permutations of the $n$ distinct numbers. Let $t_n$ be the average times of comparisons during the execution of quicksort (the average is taken over all inputs).

    Find the general expression for $t_n$. (In the problem, we do care about the constant appearing in the main term. As a result, your expression should be of the form $t_n=f(n)(1+o(1))$ for some $f(n)$).

    \subsection*{Ground Truth:}

    $ t_n =  2n\log n(1+o(1)) $

    \subsection*{Model Prediction~(OpenAI o3): }

    $ t_n =  2n\ln n(1+o(1))  $

    \subsection*{Judgement by Qwen3-235B-A22B: }
    \paragraph{label: }
    \textbf{B} (\textcolor{red}{\ding{56}})
    \paragraph{explanation: }
    The candidate's answer uses natural logarithm (ln) while the standard answer uses log (typically base 2 in algorithm analysis). \textcolor{red}{The expressions differ by a constant factor ($\ln n$ vs. $\log n$), which violates the requirement to match the constant in the main term.} The standard answer requires the form $2n\log n(1+o(1))$, but the candidate's answer $2n\ln n(1+o(1))$ is not semantically equivalent due to the logarithm base discrepancy.

    \subsection*{Judgement by OpenAI o4-mini: }
    \paragraph{label: }
    \textbf{A} (\textcolor{green}{\ding{52}})
    \paragraph{explanation: }
    The candidate's expression $t_n = 2(n+1)H_n - 4n = 2n \ln n + O(n) = 2n \ln n(1+o(1))$ is asymptotically equivalent to the standard $2n \log n(1+o(1))$.

\end{case}

\begin{case}{Case in Materials Sci. subject}{case_materials}
    \subsection*{Question: }

    Does the recovery of the material include a decrease in dislocation density and a reduction in grain boundary area?

    \subsection*{Ground Truth:}

    The reduction in dislocation density belongs to recovery, while the decrease in grain boundary area does not belong to recovery.

    \subsection*{Model Prediction~(OpenAI o3): }

    yes, no

    \subsection*{Judgement by Qwen3-235B-A22B: }

    \paragraph{label: }
    \textbf{C} (\textcolor{red}{\ding{56}})
    \paragraph{explanation: }
    The candidate's answer contains multiple conflicting selections (`yes' and `no'), \textcolor{red}{making it invalid under the `REPETITIVE' criterion as it includes contradictory outputs in a single response.}

    \subsection*{Judgement by OpenAI o4-mini: }

    \paragraph{label: }
    \textbf{A} (\textcolor{green}{\ding{52}})
    \paragraph{explanation: }
    The candidate correctly answered `yes' for dislocation density reduction and `no' for grain boundary area reduction, matching the standard answer.

\end{case}

\begin{case}{Case in Physics subject}{case_physics}
    \subsection*{Question: }

    A bowl is symmetrically inverted and immersed in a dense fluid with a specific gravity \(SG = 15.6\). The immersion depth is \(200\,\text{mm}\) (measured along the centerline from the edge of the bowl). The height of the bowl is \(80\,\text{mm}\), and the fluid rises \(20\,\text{mm}\) inside the bowl. The bowl is made of a clay formulation with a specific gravity \(SG = 6.1\), an inner diameter of \(D = 100\,\text{mm}\), and the volume of the bowl itself is approximately \(0.9\ L\). Calculate the force \(F\) required to hold the bowl in its current position.

    \subsection*{Ground Truth:}

    $F=159.4\text{N}$

    \subsection*{Model Prediction~(OpenAI o3): }

    $1.6 \times 10^2 \text{N (downward)}$

    \subsection*{Judgement by Qwen3-235B-A22B: }

    \paragraph{label: }
    \textbf{B} (\textcolor{red}{\ding{56}})
    \paragraph{explanation: }
    \textcolor{red}{The candidate's answer $1.6 \times 10^2\text{N} = 160\text{N}$ differs from the standard answer $159.4\text{N}$ by $0.6\text{N}$, exceeding the ±0.1 tolerance for numerical equivalence.} While the direction (downward) is physically reasonable, the standard answer does not include directional information, and the magnitude discrepancy invalidates strict equivalence.

    \subsection*{Judgement by OpenAI o4-mini: }

    \paragraph{label: }
    \textbf{A} (\textcolor{green}{\ding{52}})
    \paragraph{explanation: }
    The candidate's value of $1.6 \times 10^2 \text{N} (160 \text{N})$ is within the acceptable tolerance of the standard $159.4 \text{N}$. The added direction does not conflict with the standard.

\end{case}
\section{Discussion and Future Work}

\subsection{Implications of the Results}
The benchmark results from \name clearly indicate that while large models exhibit astonishing capabilities in many areas, there remains a huge gap between them and human experts in scientific domains that require deep, rigorous, and comprehensive reasoning. This finding is significant for our understanding and planning of the path toward Artificial General Intelligence (AGI). It suggests that true general intelligence lies not only in linguistic fluency and breadth of knowledge but, more importantly, in mastering the structured, verifiable, and highly complex reasoning paradigm of science. The evaluation results of \name provide us with a sober and quantitative measure of how far we are from achieving AI capable of reliable scientific discovery. This aligns with the vision of projects like HLE, which aim to provide a ``roadmap'' for future research \citep{yue2025hle}.

\subsection{Limitations of \name}
We also recognize the limitations of our current work. First, the scale of the initial dataset, with about 800 problems, while involving a huge investment in the quality of each problem, is smaller in total number than some larger-scale benchmarks. Second, the current version of \name is predominantly composed of Chinese and seven core scientific subjects. These limitations are the starting point for our future work and also reflect our commitment to academic rigor.

\subsection{The \name Platform: A Future Roadmap}
The \name project is not a one-time data release but the beginning of a long-term, continuous construction plan. Our vision is to build an open, collaborative \name platform to continuously promote the development of AI scientific reasoning capabilities. The future roadmap includes:
\begin{itemize}
    \item \textbf{Continuous Content Updates:} We plan to regularly release new problem packs to keep pace with the rapid development of models and prevent "overfitting" of the evaluation benchmark. This will ensure that \name can serve as an effective tool for measuring the capabilities of frontier models over the long term.
    \item \textbf{Expanding the Scope of Evaluation:} Future versions will gradually expand their coverage to include more scientific fields (such as neuroscience, pharmacy, environmental science, etc.), more languages (especially English), and possibly new task formats (such as hypothesis generation, experimental design, literature review, etc.), to more comprehensively evaluate the scientific capabilities of models.
    \item \textbf{Building a Community Collaboration Ecosystem:} We will establish a collaborative platform to invite domain experts from around the world to participate in the problem creation and review process. By drawing on the community collaboration models of projects like HLE \citep{yue2025hle}, we can gather a broader range of wisdom to ensure the continuous high-quality development of \name and make it a public resource truly owned and maintained by both the scientific and AI communities.
\end{itemize}

\section{Conclusion}
In response to the challenges of benchmark saturation and data contamination faced by current large model evaluations, this paper proposes and constructs a new high-difficulty, multidisciplinary scientific reasoning benchmark—\name. Through a rigorous process combining expert-original problem creation, adversarial model filtering, and multi-stage blind review, we have ensured the high standards of originality, quality, and difficulty of \name. The initial dataset focuses on the core areas of AI for Science, presented in Chinese, filling an important gap in the existing evaluation ecosystem.

Systematic evaluation of the most advanced current models shows that all models perform poorly on \name, confirming the benchmark's effectiveness as a ``touchstone'' for frontier capabilities and revealing significant deficiencies in the deep scientific reasoning of the current AI. Through in-depth analysis of model performance and error patterns, we provide specific diagnostic information and research directions for future AI model improvements.

We believe that \name and its future continuous development will provide a valuable and reliable tool for measuring and guiding the progress of AI capabilities in the key area of scientific discovery, thereby promoting the arrival of Artificial General Intelligence more robustly and clearly.

\clearpage
\bibliographystyle{plainnat}
\bibliography{ref}

\clearpage
\appendix
\newtcolorbox[auto counter, number within=section]{casebox}[2][]{%
  colback=white, 
  colframe=blue!40!black, 
  width=\textwidth,
  arc=2mm, 
  boxrule=0.5mm, 
  title={\normalsize\faCompass\hspace{0.5em}#2},
  breakable, 
  fonttitle=\bfseries\Large, 
  fontupper=\small
  #1
}

\section{\name Details}

\subsection{\name Subjects}
We show all the subjects and sub-fields of \name in \Cref{tab:subfield_stats}.
\begin{table}[t]
  \centering
  \caption{Statistics of the \name dataset, detailing the number of questions per sub-field.}
  \label{tab:subfield_stats}
  \resizebox{0.73\textwidth}{!}{
  \begin{tabular}{llr}
    \toprule
    \textbf{Subject} & \textbf{Sub-field} & \textbf{Count} \\
    \midrule
    \multirow{8}{*}{\textbf{Biology}} &
    Molecular Biology and Biotechnology & 11 \\
    & Genetics and Bioinformatics & 59 \\
    & Immunology & 2 \\
    & Physiology and Integrative Biology & 3 \\
    & Neuroscience and Psychology & 3 \\
    & Ecology & 5 \\
    & Biophysics and Biochemistry & 15 \\
    & Cell Biology & 4 \\
    \midrule
    \multirow{6}{*}{\textbf{Mathematics}} &
    Analysis & 18 \\
    & Statistics and Operations Research & 9 \\
    & Algebra and Geometry & 51 \\
    & Differential Equations and Dynamical Systems & 9 \\
    & Computational Mathematics & 3 \\
    & Interdisciplinary Mathematics & 4 \\
    \midrule
    \multirow{6}{*}{\textbf{Chemistry}} &
    Physical Chemistry & 21 \\
    & Inorganic Chemistry & 69 \\
    & Organic Chemistry & 8 \\
    & Analytical Chemistry & 11 \\
    & Chemical Engineering and Technology & 2 \\
    & Theoretical and Computational Chemistry & 6 \\
    \midrule
    \multirow{7}{*}{\textbf{Physics}} &
    Relativity & 11 \\
    & Astrophysics & 5 \\
    & Thermodynamics and Statistical Physics & 22 \\
    & Electrodynamics & 50 \\
    & Quantum Mechanics & 33 \\
    & Classical Mechanics & 48 \\
    & Fluid Mechanics & 6 \\
    \midrule
    \multirow{4}{*}{\textbf{CS}} &
    Computer Science and Technology Fundamentals & 15 \\
    & Computer Architecture & 27 \\
    & Artificial Intelligence & 16 \\
    & Computer Software & 3 \\
    \midrule
    \multirow{9}{*}{\textbf{Earth Sci.}} &
    Geography & 19 \\
    & Geodesy & 19 \\
    & Space Physics & 9 \\
    & Atmospheric Chemistry & 31 \\
    & Solid Earth Geophysics & 7 \\
    & Marine Science & 5 \\
    & Hydrology & 11 \\
    & Geochemistry & 3 \\
    & Geology & 5 \\
    \midrule
    \multirow{7}{*}{\textbf{Materials Sci.}} &
    Material Synthesis and Processing Technology & 15 \\
    & Metal Materials & 13 \\
    & Fundamental Materials Science & 7 \\
    & Material Testing and Analysis Technology & 11 \\
    & Composite Materials & 64 \\
    & Organic Polymer Materials & 23 \\
    & Materials Surface and Interface & 7 \\
    \midrule
    \multicolumn{2}{r}{\textbf{Total}} & \textbf{798} \\
    \bottomrule
  \end{tabular}}
\end{table}

\subsection{\name Question Type Distribution}
\begin{figure}[ht!]
    \centering
    \includegraphics[width=0.6\linewidth]{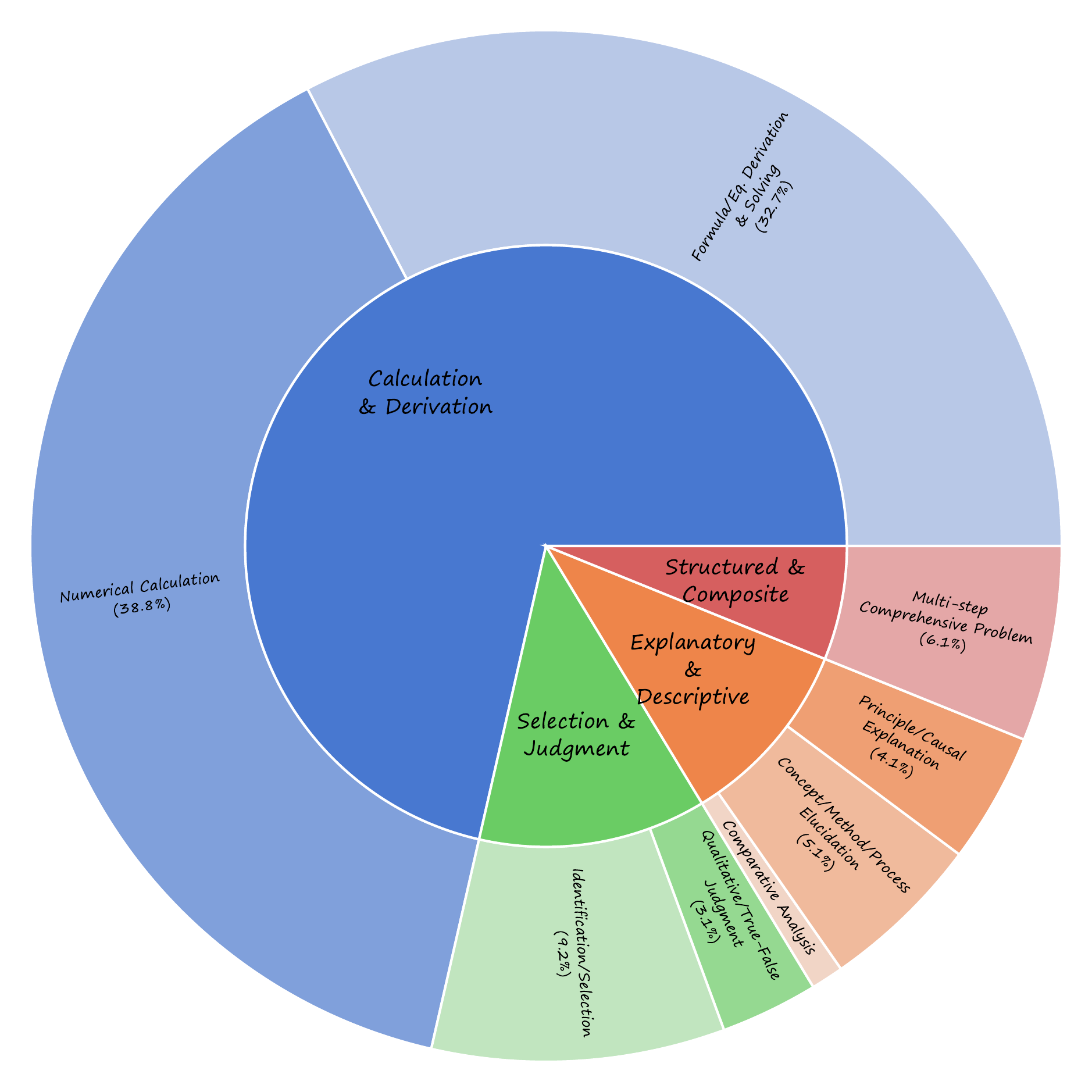}
    \caption{Hierarchical Distribution of Question Types in \name}
    \label{fig:question_type_map}
\end{figure}

We also analyze the question type of \name in \Cref{fig:question_type_map} with the definition in \Cref{tab:question_type_define}.
\begin{table}[h!]
\centering
\caption{A Typology of Question Forms Based on Structural Analysis. This table categorizes questions not by their subject matter, but by their formal structure and expected answer type.}
\label{tab:question_type_define}
\begin{tabular}{@{}lp{4cm}p{6cm}@{}}
\toprule
\textbf{Primary Category} & \textbf{Secondary Category} & \textbf{Features \& Description} \\
\midrule

\multirow{2}{*}{Calculation \& Derivation} 
& Single-Value Calculation & Requires calculating a specific numerical result based on given conditions and data. \\
\cmidrule(l){2-3}
& Formula Derivation & Requires deriving the mathematical relationship between variables; the answer is a formula or equation. \\
\midrule

\multirow{4}{*}{Explanatory \& Descriptive} 
& Principle/Causal Explanation & Typically asks "Why," requiring an explanation of the scientific principles behind a phenomenon or conclusion. \\
\cmidrule(l){2-3}
& Process/Method Description & Typically asks "How," requiring a description of operational steps or mechanisms. \\
\cmidrule(l){2-3}
& Concept/Feature Elucidation & Requires defining terms, or listing and describing the types and features of an object or concept. \\
\cmidrule(l){2-3}
& Comparative Analysis & Requires comparing the similarities and differences between two or more items. \\
\midrule

\multirow{2}{*}{Selection \& Judgment} 
& Multiple Choice / Select & Presents multiple statements, requiring the selection of all correct or qualifying items. \\
\cmidrule(l){2-3}
& True/False Judgment & Presents a single statement, requiring a judgment of its correctness (true or false). \\
\midrule

\multirow{2}{*}{Specific Info. Retrieval} 
& Fill-in-the-Blank (Cloze) & A descriptive text with blanks that need to be filled with the correct technical terms or information. \\
\cmidrule(l){2-3}
& Direct Information Recall & Requires writing specific information like chemical formulas or equations directly from the prompt. \\
\midrule

Structured Multi-Part Problem 
& Comprehensive Analysis & Revolves around a central scenario, with multiple, logically connected sub-questions. \\
\bottomrule
\end{tabular}
\end{table}

\section{Question Contributors}
\begin{figure}[ht!]
    \centering
    \includegraphics[width=0.8\linewidth]{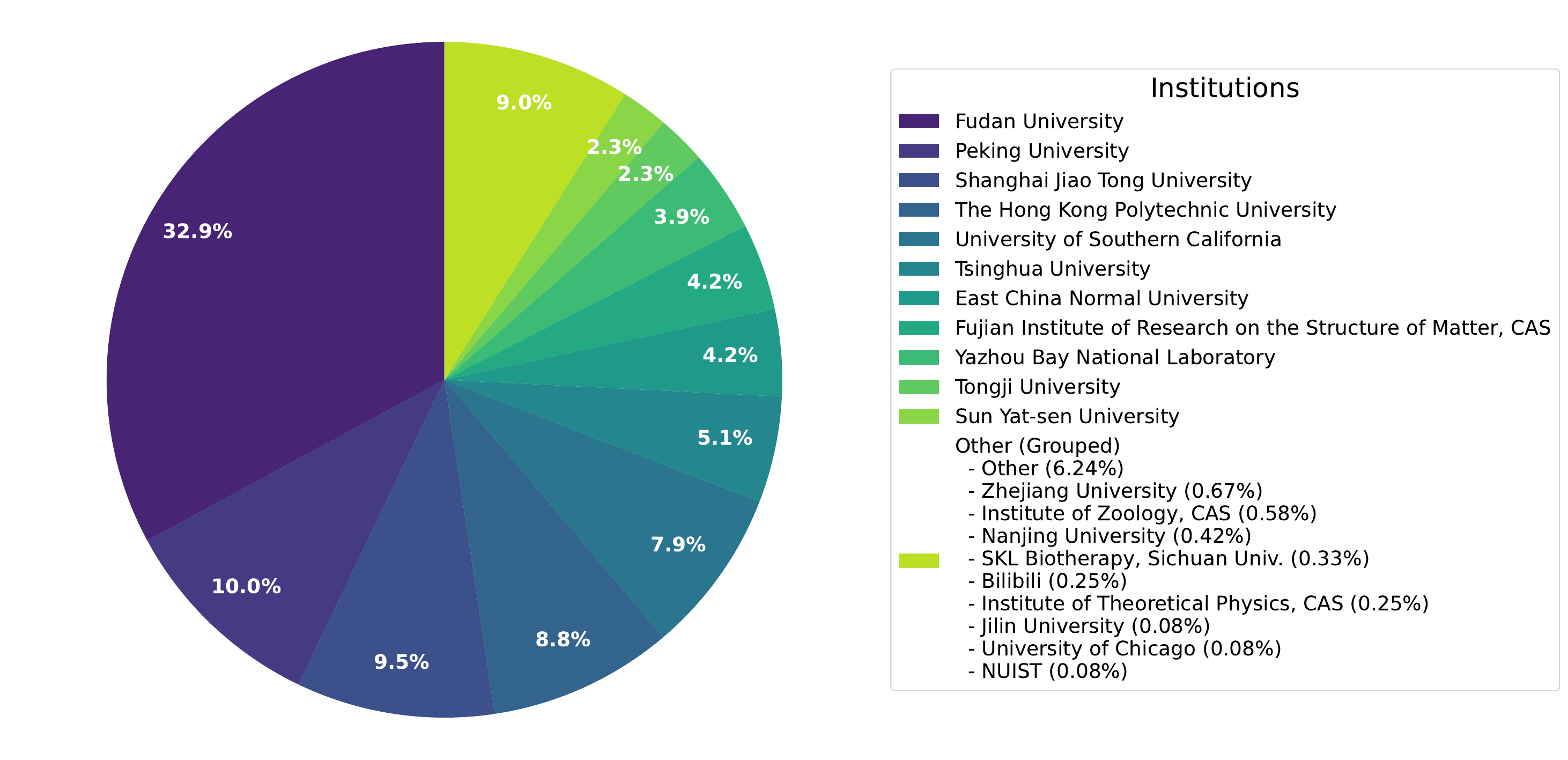}
    \caption{Distribution of Question Contributing Institutions.}
    \label{fig:question_contributing_institutions}
\end{figure}

We have collaborated with scholars from over 25 different universities or research organizations to contribute to \name, and the statistics of these institutions are shown in \Cref{fig:question_contributing_institutions}.

\section{Expert Review}
\label{expert_review}
\subsection{Peer Review Stage}
\begin{table}[h]
  \caption{\name Expert Peer Review Scoring Rubric (Math Domain as a Tempalte).}
  \label{tab:review_criteria}
  \centering
  \small
  \begin{tabular}{p{3cm}cp{7cm}}
    \toprule
    \textbf{Dimension} & \textbf{Score} & \textbf{Standard Description} \\
    \midrule
    \textbf{Content \& Format} & 5 & Content is rigorous, format is perfect, expression is precise. \\
    (0-5) & 3 & Content is reasonable, with minor areas for improvement. \\
    & 1 & Multiple clear errors, requires major revision. \\
    \midrule
    \textbf{Scientific Value} & 5 & Extremely high value, tests cross-domain integration and top-tier thinking. \\
    (0-5) & 3 & High value, tests application of multiple knowledge points. \\
    & 1 & Limited value, tests simple memorization or mechanical operation. \\
    \midrule
    \textbf{Difficulty Rating} & 5 & Extremely difficult (e.g., IMO/Putnam level for Math). \\
    (0-5) & 3 & High difficulty (e.g., AIME/CMO level for Math). \\
    & 1 & Basic level (e.g., High school foundation). \\
    \bottomrule
  \end{tabular}
\end{table}
We employ a structured peer review process governed by a formal scoring rubric, exemplified in \Cref{tab:review_criteria} for the mathematics domain. Each problem is independently evaluated by two to three domain experts who score its quality across three core dimensions: (1) Content \& Format, (2) Scientific Value, and (3) Difficulty. A problem is advanced from the peer review stage only if it achieves an average score of at least 3.0 across all reviews.

\subsection{Meta Review Stage}
\begin{table}[h]
  \caption{Analysis of Rejection Reasons for Meta Question Review}
  \label{tab:meta_error_analysis}
  \centering
  \begin{tabular}{llr}
    \toprule
    \textbf{Main Category} & \textbf{Subcategory} & \textbf{Percentage} \\
    \midrule
    \textbf{Content \& Logical Flaws} & \textit{\textbf{Subtotal}} & \textbf{46\%} \\
    & Incorrect Answer or Fact & 16\% \\
    & Calculation or Derivation Error & 14\% \\
    & Oversimplified or Missing Premise & 8\% \\
    & Flawed Logic / Violates Principles & 6\% \\
    & Ignores Provided Context/Data & 2\% \\
    \midrule
    \textbf{Difficulty \& Scope} & \textit{\textbf{Subtotal}} & \textbf{38\%} \\
    & Difficulty Too Low & 24\% \\
    & Out of Scope / Uncollected Type & 8\% \\
    & Low Value / Rote Memorization & 6\% \\
    \midrule
    \textbf{Content Quality \& Formatting} & \textit{\textbf{Subtotal}} & \textbf{16\%} \\
    & Formatting or Rendering Error & 8\% \\
    & Missing or Unclear Content & 4\% \\
    & Mismatched or Hard-to-Verify Content & 4\% \\
    \bottomrule
  \end{tabular}
\end{table}
The meta review stage we invite experts to give a "Meta Review" for every question passed in peer review stage, we sample 50 questions failed in this stage and summarize the reason in \Cref{tab:meta_error_analysis}.

\section{Expert Review}

\section{Prompts for Evaluation}

Prompt \ref{pp:prompt_inference} and Prompt \ref{pp:prompt_judgment} demonstrates the details instructions leveraged for evaluation.

\clearpage
\begin{prompt}{Prompt for Prediction}{prompt_inference}
\subsection*{Problem:}
\begin{lstlisting}
{problem}
\end{lstlisting}

\subsection*{Instructions:}
\begin{itemize}
\item Solve the problem step by step. If the problem contains multiple sub-questions, make sure to solve each one individually.
\item At the end, output \textbf{only} the final answers in the following format:
\end{itemize}

\begin{lstlisting}
```json
{
  "answers": [
    "answer to sub-question 1", 
    "answer to sub-question 2",
    ...
  ]
}
```
\end{lstlisting}

\begin{itemize}
\item Each item in the list should be the \textbf{final answer} to a sub-question.
\item If there is only one question, return a list with a single item.
\item \textbf{Do not} include any explanation, reasoning steps, or additional text outside the JSON list.
\item \textbf{Do} put the JSON list in the block of \verb|```json ... ```|
\end{itemize}
\end{prompt}

\begin{prompt}{Prompt for Judgement}{prompt_judgment}
  You are an expert answer grader. Your task is to evaluate whether the
  candidate's \textbf{final answer} matches the \textbf{provided standard
  answer}. Follow the grading protocol strictly and \textbf{do not
  generate or modify answers}. Only compare the candidate's response to
  the given standard.
  
  \begin{center}\rule{0.5\linewidth}{0.5pt}\end{center}
  
  \subsection*{Evaluation Guidelines}
  \paragraph{1. Reference Standard}
  
  \begin{itemize}
  \tightlist
  \item
    The \textbf{standard answer is always correct} --- never question its
    validity.
  \item
    The \textbf{question itself is valid} --- do not critique or
    reinterpret it.
  \item
    Do \textbf{not} regenerate, fix, or complete the candidate's answer
    --- only \textbf{evaluate} what is provided.
  \end{itemize}
  
  \paragraph{2. Comparison Strategy}
  
  \begin{itemize}
  \item
    Carefully analyze the \textbf{question type} and \textbf{standard
    answer format}:
  
    \begin{itemize}
    \tightlist
    \item
      Determine whether an \textbf{exact match} is required, or whether
      \textbf{partial correctness} is acceptable (e.g., for
      multi-component or expression-based answers).
    \item
      This judgment should be based on the \textbf{question's phrasing and
      answer structure}.
    \end{itemize}
  \item
    Evaluate \textbf{only the candidate's final answer}, ignoring
    reasoning or explanation.
  \item
    Ignore differences in \textbf{formatting, style}, or \textbf{variable
    naming}, as long as the content is equivalent.
  \item
    For \textbf{mathematical expressions}, check \textbf{step-by-step
    equivalence} (e.g., by simplifying both expressions and comparing
    results).
  \item
    For \textbf{multiple-choice questions}, only the \textbf{final
    selected option} and its \textbf{associated content} matter.
  \item
    For \textbf{decimal or fraction comparisons}, consider the answers
    equivalent if the relative error is \textbf{$\le$ ±0.1}.
  \end{itemize}
  
  \paragraph{3. Multi-part Answers}
  
  \begin{itemize}
  \tightlist
  \item
    If the question requires \textbf{multiple components or selections},
    all parts must match the standard answer exactly.
  \item
    Compare each component individually.
  \item
    \textbf{Partial correctness is not acceptable} --- label as incorrect
    if any part is wrong.
  \end{itemize}
  
  \paragraph{4. Validity Check}
  
  Immediately reject the candidate's answer if it meets \textbf{any of the
  following criteria}:
  
  \begin{itemize}
  \tightlist
  \item
    \textbf{INCOMPLETE}: Final sentence is cut off or the answer is
    clearly unfinished.
  \item
    \textbf{REPETITIVE}: Contains repeated phrases or outputs in a loop.
  \item
    \textbf{REFUSAL}: Explicitly states inability to answer (e.g., ``I
    cannot answer this question'').
  \item
    Use label \textbf{C}.
  \end{itemize}
  
  \begin{center}\rule{0.5\linewidth}{0.5pt}\end{center}
  
  \subsection*{Grading Scale}
  
  \begin{longtable}{ll>{\raggedright\arraybackslash}p{0.7\textwidth}}
  \toprule
  Grade & Label & Description \\
  \midrule
  A & CORRECT & Exact or semantically equivalent match; includes numerically equivalent results (within ±0.0001) \\
  B & INCORRECT & Any deviation from the standard answer; includes partial matches \\
  C & INVALID & Answer is INCOMPLETE, REPETITIVE, or a REFUSAL \\
  \bottomrule
  \end{longtable}
  
  \begin{center}\rule{0.5\linewidth}{0.5pt}\end{center}
  
  \subsection*{Evaluation Procedure \& Output
  Format}
  
  \begin{enumerate}
  \def\labelenumi{\arabic{enumi}.}
  \item
    \textbf{Check for Validity First}:
  
    \begin{itemize}
    \tightlist
    \item
      If the answer is incomplete, repetitive, or a refusal,
      \textbf{immediately assign label C} with the reason and stop further
      evaluation.
    \end{itemize}
  \item
    \textbf{If Valid, Compare Content}:
  
    \begin{itemize}
    \item
      Analyze the question type: Are strict matches required (e.g., order,
      format, completeness)?
    \item
      Apply tolerances: Accept allowed variations (e.g., unformatted but
      equivalent math, missing labels in MCQs).
    \item
      Carefully compare final answers for:
  
      \begin{itemize}
      \tightlist
      \item
        Semantic or mathematical equivalence
      \item
        Relative error tolerance (±0.1)
      \item
        Expression format flexibility
      \end{itemize}
    \end{itemize}
  \item
    \textbf{Produce a Final Judgment}:
  
    \begin{itemize}
    \item
      For each sub-question, return:
  
      \begin{lstlisting}
```json
{
  "label": "A" / "B" / "C",
  "explanation": "Brief justification here"
}
```
      \end{lstlisting}
    \item
      At the end, return a list of these JSON objects for each
      sub-question.
  
      \begin{lstlisting}
```json
{
  "judgements": [
    {
        "label": "A" / "B" / "C" for sub-question 1,
        "explanation": "Brief justification here for sub-question 1"
    },
    {
        "label": "A" / "B" / "C" for sub-question 2,
        "explanation": "Brief justification here for sub-question 2"
    },
    ...
  ]
}
```
      \end{lstlisting}
    \item
      If there is only one question, return a list with a single item.
    \item
      \textbf{Do} put the JSON list in the block of \texttt{json}.
    \end{itemize}
  \end{enumerate}
  
  \begin{center}\rule{0.5\linewidth}{0.5pt}\end{center}
  
  \subsection*{Task Input}
  
  \begin{lstlisting}
  
  {problem}
  
  
  
  {answer}
  
  
  
  {prediction}
  
  \end{lstlisting}
  
  \begin{center}\rule{0.5\linewidth}{0.5pt}\end{center}
  
  \subsection*{Begin Evaluation Below:}
  
  Analyze the candidate's answer step by step, then provide a
  \textbf{final structured judgment}.
\end{prompt}
\clearpage

\section{Performance on the Test Set of \name} \label{app:test_performance}

\paragraph{Overall Performance. }
\Cref{tab:test_performance_judge_oss} presents the performance of all LLMs evaluated on the test set of \name, as judged by GPT-OSS-120B, and ordered by average accuracy.
OpenAI GPT-5-High ranks highest with an accuracy of 43.8\%, followed by Gemini-2.5-Pro at 39.9\%, OpenAI o3-High at 37.4\%, and Grok-4 at 35.4\%. 
The lower-performing models include Qwen3-235B-A22B-2507 at 39.6\%, Doubao-Seed-1.6-thinking at 28.8\%, DeepSeek-R1-0528 at 26.1\%, OpenAI o4-mini at 24.1\%, and GPT-OSS-120B at 23.3\%. 
The mG-Pass@2 and mG-Pass@4 scores, which indicate stability across multiple predictions, exhibit a similar pattern, with OpenAI GPT-5-High achieving the highest scores of 34.2\% and 33.5\%, respectively, while GPT-OSS-120B scores the lowest, at 14.6\% and 12.8\%.
In comparison to \Cref{tab:model_performance_judge_oss}, where OpenAI GPT-5-High leads with an accuracy of 42.9\%, followed closely by Gemini-2.5-Pro at 35.3\%, the overall ranking remains consistent, though OpenAI o3 models show competitive performance in the mid range.
Furthermore, the accuracy of OpenAI o4-mini shows only a slight variation, from 24.1\% in \Cref{tab:model_performance_judge_oss} to 22.4\% in \Cref{tab:test_performance_judge_oss}, suggesting relative consistency. Other models also demonstrate minor fluctuations.

\begin{table}[t]
    \centering
    \caption{The performance of various LLMs on the test set of \name, as judged by GPT-OSS-120B, is sorted by average accuracy. Each LLM is prompted to generate four predictions, and we report the average accuracy as well as the mG-Pass@$\{2,4\}$ scores. A high mG-Pass score indicates a high level of stability across multiple predictions.} \label{tab:test_performance_judge_oss}
    \resizebox{\textwidth}{!}{
        \begin{tabular}{lcrrr}
            \toprule
            \textbf{Model} & \textbf{\#Tokens} & \textbf{Accuracy (\%)} $\uparrow$ & \textbf{mG-Pass@2 (\%)} $\uparrow$ & \textbf{mG-Pass@4 (\%)} $\uparrow$ \\
            \midrule
            OpenAI GPT-5-High & 32k & 43.8 & 34.2 & 33.5 \\
            Gemini-2.5-Pro & 32k & 39.9 & 30.6 & 28.8 \\
            OpenAI o3-High & 32k & 37.4 & 25.9 & 23.8 \\
            Grok-4 & 32k & 35.4 & 26.2 & 24.4 \\
            Qwen3-235B-A22B-2507 & 32k & 29.7 & 21.6 & 19.8 \\
            DeepSeek-V3.1 & 32k & 29.5 & 20.2 & 18.5 \\
            Doubao-Seed-1.6-thinking & 32k & 28.8 & 20.1 & 18.5 \\
            DeepSeek-R1-0528 & 32k & 26.1 & 18.4 & 16.6 \\
            OpenAI o4-mini & 32k & 24.1 & 15.2 & 13.6 \\  
            GPT-OSS-120B-High & 32k & 23.3 & 14.6 & 12.8 \\
            \bottomrule
        \end{tabular}
    }
\end{table}

\begin{figure}[t!]
  \centering
  \includegraphics[width=\linewidth]{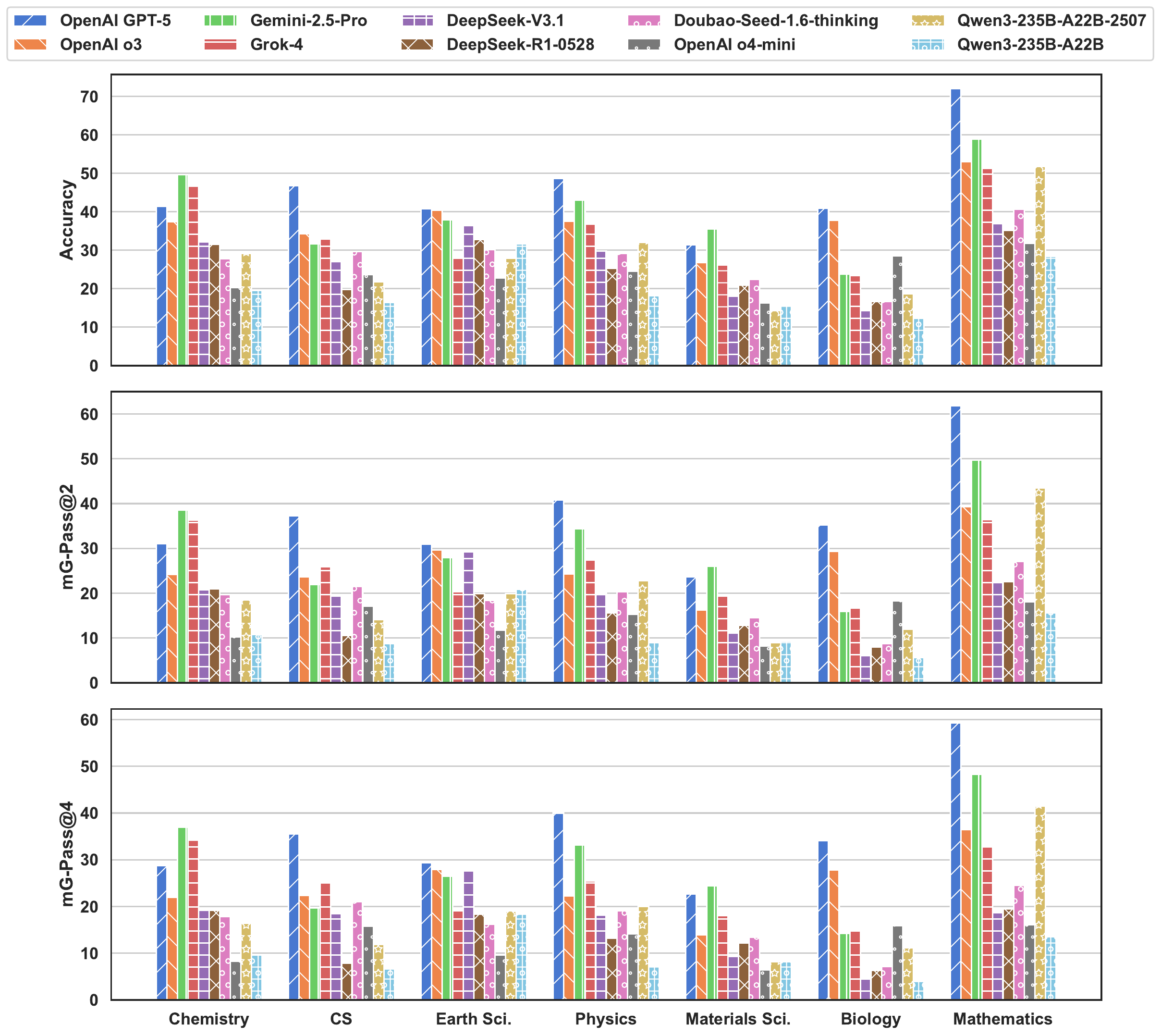}
  \caption{The performance of different LLMs across different subjects of \name's test set.} \label{fig:test_subject_performance}
\end{figure}

\paragraph{Subject Performance. }
\Cref{fig:test_subject_performance} illustrates the performance of all LLMs across different subjects in \name's test set. 
OpenAI GPT-5 consistently achieves the highest accuracy and mG-Pass scores across all subjects, standing out as the clear leader. 
Gemini-2.5-Pro also delivers competitive results, particularly in Chemistry, Physics, and Biology. 
Grok-4 demonstrates notable strength in Computer Science, achieving the best scores in this domain. 
In contrast, Qwen3-235B-A22B and Qwen3-235B-A22B-2507 generally show weaker performance across most subjects, while DeepSeek-R1-0528 and OpenAI o4-mini remain in the lower tier with moderate results. 
Doubao-Seed-1.6-thinking and DeepSeek-V3.1 produce mixed outcomes, performing well in some subjects but lagging in others.  
For Specific Subjects: 
\begin{itemize}[leftmargin=*] 
  \item \textbf{Chemistry}: OpenAI GPT-5 leads by a large margin, followed by Gemini-2.5-Pro, with Grok-4 and Doubao-Seed-1.6-thinking showing moderate results. 
  \item \textbf{Computer Science}: Grok-4 achieves the best overall performance, with GPT-5, o3, and Doubao-Seed-1.6-thinking trailing behind. 
  \item \textbf{Earth Science}: GPT-5 ranks highest, while Gemini-2.5-Pro and o3 achieve competitive performance. 
  \item \textbf{Physics}: GPT-5 dominates, with Gemini-2.5-Pro also performing strongly. 
  \item \textbf{Materials Science}: GPT-5 again leads, followed by Gemini-2.5-Pro and o3 as the next tier of models. 
  \item \textbf{Biology}: GPT-5 significantly surpasses all other models, while Gemini-2.5-Pro and o3 achieve moderate accuracy and stability. 
  \item \textbf{Mathematics}: GPT-5 shows overwhelming dominance, with Qwen3-235B-A22B-2507 and Gemini-2.5-Pro forming the second tier of performance.  
\end{itemize}
By comparing these performances with those on the validation set~(\Cref{fig:test_subject_performance}), we can assess the consistency of the subject-specific outcomes. 
For example, GPT-5 consistently dominates across both datasets, while Grok-4 maintains its strength in Computer Science. 
Such consistency highlights the inherent strengths and weaknesses of the models across knowledge domains.

\section{Contributors}
\label{app:contributors}

Our team is composed of researchers with diverse technical backgrounds, each of whom contributed in different ways to the success of this project. The core contributors were responsible for all stages of the work, including data collection strategy, data quality screening, evaluation design, result analysis, and manuscript preparation. Project contributors, who come from multiple research communities, coordinated data collection efforts and oversaw data quality control. Data contributors provided realistic and challenging questions, bringing their domain expertise to strengthen the dataset. The corresponding authors initiated and supervised the project and secured the resources necessary to complete this work.







\begin{center}
\begin{longtable}{p{0.25\textwidth} p{0.7\textwidth}}
\caption{List of Contributors}\\
\hline
\textbf{Contribution Type} & \textbf{Contributors} \\
\hline
\endfirsthead
\multicolumn{2}{c}%
{{\tablename\ \thetable{} -- continued from previous page}} \\
\hline
\textbf{Contribution Type} & \textbf{Contributors} \\
\hline
\endhead
\hline \multicolumn{2}{r}{{Continued on next page}} \\
\endfoot
\endlastfoot

\textbf{Core Contributor} & 
Hongwei Liu$^{1}$, 
Junnan Liu$^{1}$, 
Shudong Liu$^{1}$ \\
\hline

\textbf{Project Contributor} & 
Haodong Duan$^{1}$, 
Yuqiang Li$^{1}$, 
Mao Su$^{1}$, 
Xiaohong Liu$^{2}$, 
Guangtao Zhai$^{2}$, 
Xinyu Fang$^{3, 1}$, 
Qianhong Ma$^{2, 1}$, 
Taolin Zhang$^{4, 1}$, 
Zihan Ma$^{5, 1}$, 
Yufeng Zhao$^{4, 1}$, 
Peiheng Zhou$^{1}$, 
Linchen Xiao$^{1}$, 
Wenlong Zhang$^{1}$, 
Shijie Zhou$^{6}$, 
Xingjian Ma$^{6}$, 
Siqi Sun$^{6}$, 
Jiaye Ge$^{1}$, 
Meng Li$^{1}$, 
Yuhong Liu$^{1}$, 
Jianxin Dong$^{1}$,
Jiaying Li$^{1}$, 
Hui Wu$^{1}$, 
Hanwen Liang$^{1}$,
Jintai Lin$^{15}$,
Yanting Wang$^{17}$,
Jie Dong$^{2}$,
Tong Zhu$^{16}$,
Tianfan Fu$^{20}$,
Conghui He$^{1}$,
Qi Zhang$^{6}$\\
\hline

\textbf{Corresponding Author} & 
Lei Bai$^{1}$, 
Kai Chen$^{1}$, 
Songyang Zhang$^{1}$ \\
\hline

\textbf{Data Contributors} & 
Yuqiang Li$^{2}$, 
Ben Gao$^{7}$, 
Mao Su$^{1}$, 
Shengdu Chai$^{1, 6}$, 
Xuefeng Wei$^{8}$, 
Zicheng Zhang$^{2}$, 
Chunyi Li$^{2}$, 
Yiheng Wang$^{2, 1}$, 
Weijia Li$^{9}$, 
Fenghua Ling$^{1}$
% ...
\\

\end{longtable}
\end{center}

\clearpage
\vspace{1cm}
\noindent\textbf{Main Affiliations} \\
$^{1}$ Shanghai AI Lab \\
$^{2}$ Shanghai Jiao Tong University \\
% ...
$^{6}$ Fudan University \\
$^{7}$ Wuhan University \\
$^{8}$ Fujian Institute of Research on the Structure of Matter, Chinese Academy of Sciences \\
$^{9}$ Sun Yat-sen University \\
% ...

\end{document}